\documentclass[lettersize,journal]{IEEEtran}
\usepackage{amsmath,amsfonts,amssymb}
\usepackage{array}
\usepackage{textcomp}
\usepackage{stfloats}
\usepackage{url}
\usepackage{graphicx}
\usepackage{cite}
\usepackage[T1]{fontenc}
\usepackage{multirow}
\usepackage{float}
\usepackage{xcolor}
\usepackage{mathtools}
\usepackage{hyperref}
\allowdisplaybreaks%

\usepackage{copyright}

\begin{document}

\title{A Stochastic Framework for Continuous-Time \\ State Estimation of Continuum Robots}

\author{Spencer Teetaert,~\IEEEmembership{Student Member,~IEEE,} Sven Lilge,~\IEEEmembership{Member,~IEEE,} \\Jessica Burgner-Kahrs,~\IEEEmembership{Senior Member,~IEEE,} Timothy D. Barfoot,~\IEEEmembership{Fellow,~IEEE}
\thanks{The authors are with the University of Toronto Robotics Institute, Canada (corresponding e-mail: \texttt{spencer.teetaert@robotics.utias.utoronto.ca}). This work was supported in part by the Natural Sciences and Engineering Research Council of Canada (NSERC) and the Ontario Graduate Scholarship.}
}

\markboth{IEEE Transactions on Robotics, VOL.~??, ????}%
{Teetaert \MakeLowercase{\textit{et al.}}: A Stochastic Framework for Continuous-Time State Estimation of Continuum Robots}

\copyrightnotice

\maketitle

\begin{abstract}
  State estimation techniques for continuum robots (CRs) typically involve using computationally complex dynamic models, simplistic shape approximations, or are limited to quasi-static methods. These approaches can be sensitive to unmodeled disturbances acting on the robot. Inspired by a factor-graph optimization paradigm, this work introduces a continuous-time stochastic state estimation framework for continuum robots. We introduce factors based on continuous-time kinematics that are corrupted by a white noise Gaussian process (GP). By using a simple robot model paired with high-rate sensing, we show adaptability to unmodeled external forces and data dropout. The result contains an estimate of the mean and covariance for the robot's pose, velocity, and strain, each of which can be interpolated continuously in both space and time. This same interpolation scheme can be used during estimation, allowing for inclusion of measurements that are asynchronous to the estimated states. Our method's inherent sparsity leads to a linear solve complexity with respect to time and interpolation queries in constant time. We demonstrate our method on a CR with gyroscope and pose sensors, highlighting its versatility in real-world systems.
\end{abstract}

\begin{IEEEkeywords}
  Probability and Statistical Methods, Flexible Robots, Dynamics, State Estimation
\end{IEEEkeywords}

\section{Introduction}

Continuum robots (CRs) are jointless, flexible, and easily miniaturizable manipulators capable of bending into contorted spatial shapes.
They are often said to be inspired by the animal kingdom, resembling the motion of snakes, elephant trunks, or worms~\cite{Robinson1999}.
Their unique properties allow them to navigate in confined and cluttered environments.
This makes them particularly suitable for applications such as minimally invasive surgery~\cite{Burgner-Kahrs2015,Dupont2022}, industrial inspection and repair in hard-to-reach places~\cite{Dong2017,Wang2021}, as well as search and rescue operations in disaster areas~\cite{Hawkes2017}.

\begin{figure}[t!]
    \centering
    \includegraphics[width=\columnwidth]{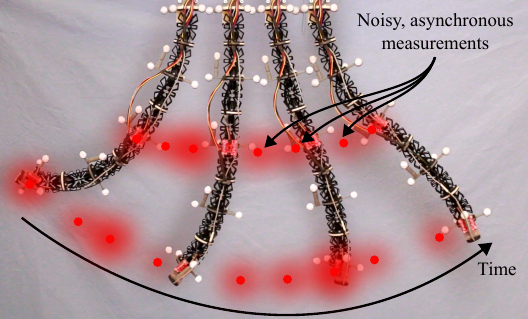}
    \caption{A CR moving through time, taking asynchronous and noisy measurements of its state. This situation arises in practical scenarios where multiple high-rate sensors are used on the robot. In such scenarios, data from multiple high-rate sensors produce update rates much faster than a CR shape can reasonably be solved for. In this work, we propose addressing this issue by using a physically motivated Gaussian process prior defined over both space and time that is capable of continuous-time interpolation in constant time. This efficient interpolation can be used to fuse high-rate measurements from multiple sensors at slower time steps while maintaining accurate estimates in between time steps.}\label{fig:title_fig}
\end{figure}

To date, great progress has been made on the modeling of continuum robots~\cite{Armanini2023}, using a variety of kinematic, static, and dynamic assumptions and approaches.
Such methods aim to accurately predict the robot's shape given its material properties, actuation, and external loads, which is crucial for the aforementioned applications. 
However, they are usually prone to inaccuracies arising from both unmodeled effects and unknown contact forces during interaction with the environment.
This makes the application of open-loop model-based control approaches~\cite{Della2023} challenging and motivates the integration of additional sensing to account for such model inaccuracies.
Here, state estimation approaches that can fuse sensor feedback on the robot's state with appropriate model prediction become particularly important. This is especially true as real world data is noisy. Fig.~\ref{fig:title_fig} illustrates such a scenario where a continuum robot (CR) is moving through time, taking asynchronous and noisy measurements of its state.

\subsection{Related Work}

In the following, we review the most relevant related work on the state estimation of continuum robots.\IEEEpubidadjcol{}
During the review, we differentiate between non-probabilistic and probabilistic methods.
Both approaches consider quantities such as actuation and measurements to compute a resulting continuum robot state. The definition of the robot state varies by work, but commonly includes the robot's pose, strain, and/or velocity, and occasionally contains higher derivatives such as acceleration or the strain gradient.
However, probabilistic approaches explicitly take into account the uncertainties of both the underlying prior model and the noisy measurements, weighing them appropriately.
This additionally allows for computing a distribution of possible robot states, where the mean is the most likely state and the covariance is its uncertainty.

\subsubsection{Non-probabilistic Methods}

Non-probabilistic estimation methods usually employ optimization approaches to fit kinematic, static, or dynamic models to discretely sensed quantities, such as pose or strain.

Examples of fitting kinematic shape representations include optimizing constant-curvature arcs~\cite{Roesthuis2014}, Bézier curves~\cite{Song2015}, or polynomial-curvature models~\cite{Kim2014,Pei2025} to minimize the deviation between the robot's centerline and the corresponding measurements.
More recently, a model-based shape reconstruction framework based on a finite-element analysis (FEA) simulation was proposed, incorporating readings of distributed strain sensors into an optimization routine~\cite{Liu2025}.
In~\cite{Stella2024}, a piecewise-constant-curvature representation is fit to orientation readings from an inertial measurement unit (IMU), while accounting for measurement drift.
By relying not only on current measurements but also incorporating the history of readings, a sliding-window estimator is presented in~\cite{Abdelaziz2023}, which fuses a series of past IMU measurements at the robot's tip with a constant-curvature model.

Instead of solely relying on kinematic representations, some approaches incorporate assumptions and knowledge about the mechanics of continuum robots by fitting appropriate static models to measurement data.
Examples include using static models based on the principle of virtual power~\cite{Rone2013} and pseudo-rigid-body assumptions~\cite{Venkiteswaran2019}.
Similarly, more recent approaches are additionally considering the dynamic mechanical properties of continuum robots, by fusing dynamic Cosserat-rod models, expressing the robot's state as a function of space and time, with sensed quantities.
For instance, the works in~\cite{Zheng2023,Zheng2024,Zheng2025} fuse velocity and load measurements at the boundaries of the robot's shape (i.e., at its tip and base) with the partial differential equations of a dynamic Cosserat-rod model.
In~\cite{Feliu2025}, a dynamic state estimation approach of soft robots is proposed, which only relies on sensing of the actuation values and known external forces.

Lastly, approaches that leverage learned models to estimate the shape of continuum robots have started to attract increasing attention.
In one example, a neural network is employed to predict control points used for a continuous shape interpolation based on strain-gauge readings~\cite{Zhao2022}.
In another example, fiber Bragg grating (FBG), orientation, and position measurements are used to estimate continuum robot shapes utilizing a physics-informed neural network~\cite{Zhan2025}.

\subsubsection{Probabilistic Methods}

Most probabilistic state estimation approaches aim to track the evolution of a chosen continuum robot state representation over time, typically using filtering techniques.
For instance, the works~\cite{Brij2010} and~\cite{Borgstadt2015} both rely on a constant-curvature representation, tracking its arc parameters over time using particle filters to estimate a catheter's tip pose or position, respectively.
Similar approaches include~\cite{Ataka2016,Chen2019a,Loo2019,Kim2021,Mehl2024,Peng2024}, all of which implement discrete-time extended or unscented Kalman filters to estimate the piecewise-constant-curvature parameters of multi-backbone, tendon-driven, or pneumatic continuum robots over time.

Instead of relying on constant-curvature representations, some approaches utilize modal basis functions, enabling higher-order and potentially more accurate descriptions of the continuum robot's shape and deformations.
In~\cite{Lobaton2013}, learned shape basis functions are chosen to represent the continuum robot's shape, tracking their coefficients with a probabilistic filter implementation for estimation over time.
The use of several polynomial shape basis function representations of different orders is proposed in~\cite{Zhang2022}, which implements a discrete-time extended Kalman filter for the shape estimation of tendon-driven continuum robots.

Finally, there exists a class of probabilistic state estimation approaches that rely on variable-curvature representations to estimate the shape of continuum robots.
A key distinction from the previously discussed methods is that these approaches are currently employed almost exclusively in a quasi-static manner.
Rather than estimating the robot's state over time, they compute a continuous estimate of its state along the robot's length at a single point in time.
This is achieved by applying probabilistic filtering or smoothing over the robot's arclength, rather than over time as in earlier examples.

One example of such an approach is presented in~\cite{Mahoney2016a}, where a Rauch-Tung-Striebel (RTS) smoother is applied over arclength to estimate the smooth, continuous state of a concentric tube robot.
The method relies on numerical integration to solve the ordinary differential equations derived from a Cosserat-rod statics model.
This approach was later extended to coupled parallel continuum robots in~\cite{Anderson2017}.
Alternatively, recent work proposes the usage of Gaussian process regression over arclength to estimate the smooth continuum robot state~\cite{Lilge2022}, re-purposing methods from the continuous-time estimation literature~\cite{Anderson2015}.
Here, a simplified static Cosserat-rod-based prior is proposed, allowing stochastic integration in closed form, enabling fast computation times during estimation.
Recent work on this approach includes an extension to handle coupled parallel continuum robots~\cite{Lilge2025a}, the incorporation of known actuation inputs~\cite{Lilge2025b}, as well as the estimation of external loads in addition to estimating the robot's shape~\cite{Ferguson2024}.

Looking at the current state of probabilistic state estimation approaches for continuum robots, it can be seen that most methods rely on simplifying kinematic representations, such as constant-curvature assumptions or polynomial shape basis functions~\cite{Ataka2016,Loo2019,Mehl2024,Lobaton2013,Zhang2022}.
These approaches typically employ filters that operate in a discrete-time manner, requiring a state estimation update whenever a measurement becomes available and restricting the estimated robot state to these discrete update times.
Variable-curvature frameworks, on the other hand, are currently limited to quasi-static scenarios~\cite{Mahoney2016,Lilge2022,Ferguson2024}.
While dynamic variable-curvature estimation methods do exist, they rely heavily on detailed modeling assumptions and prior system knowledge~\cite{Zheng2023,Zheng2024,Zheng2025,Feliu2025}.
Additionally, they are restricted to non-probabilistic formulations, limiting their ability to incorporate model and measurement uncertainties.
As such, a lack of probabilistic methods capable of estimating the smooth state of continuum robots continuously over both time and arclength, while explicitly accounting for noise and uncertainty, can be identified.

\subsection{Contributions}

\begin{table}[t]
\caption{Comparison of Continuum Robot State Estimation Approaches}\label{tab:comparison}
\centering
\resizebox{\columnwidth}{!}{
\begin{tabular}{lcccc}
\hline
Method & Estimates Uncertainty & Dynamic & Continuous-Time & Robot Model \\
\hline
Feliu-Talegon et al.~\cite{Feliu2025} & $\times$ & \checkmark{} & \checkmark{} & Physical \\
Zheng et al.~\cite{Zheng2023} & $\times$ & \checkmark{} & \checkmark{} & Physical \\
Lobaton et al.~\cite{Lobaton2013} & $\times$ & \checkmark{} & \checkmark{} & Physical \\
Ataka et al.~\cite{Ataka2016} & $\times$ & \checkmark{} & $\times$ & Geometric \\
Borgstadt et al.~\cite{Borgstadt2015} & \checkmark{} & \checkmark{} & $\times$ & Geometric \\
Lilge et al.~\cite{Lilge2022} & \checkmark{} & $\times$ & $\times$ & Physical \\
Ferguson et al.~\cite{Ferguson2024} & \checkmark{} & $\times$ & $\times$ & Physical \\
\textbf{Ours} & \checkmark{} & \checkmark{} & \checkmark{} & Physical \\
\hline
\end{tabular}
}
\end{table}

In this work, we address this gap by proposing a physically motivated Gaussian-process prior defined over both space and time, extending concepts from continuous-time~\cite{Anderson2015} and continuous-arclength~\cite{Lilge2022} estimation into a single framework. This work enables continuous estimation of the robot's state over both time and arclength. Preliminary work aiming to address this gap was proposed in~\cite{Teetaert2024}. However, the work remains mostly incomplete and unvalidated.

To the best of our knowledge, this work represents the first state estimator for continuum robots that estimates uncertainties while being continuous in time and operating with asynchronous sensor inputs. Table~\ref{tab:comparison} highlights our work within the context of existing literature. The proposed method supports efficient handling of high-rate and asynchronous measurements with minimal computational overhead. Unlike existing approaches for CR state estimation, we do not need to solve for the full robot state at each measurement time, leading to significant computational gains. We validate our approach through both simulation and experimental evaluation, using pose and velocity measurements on a tendon-driven continuum robot. Our experimental results highlight the method's ability to capture dynamic effects in the robot's motion, handle data dropout, and benefit from sensor fusion. An open-source implementation of this project built on GTSAM~\cite{Dellaert2026} is made publicly available to the research community at \url{https://github.com/utiasASRL/space_time_continuum}. 

\section{Methodology}

\subsection{Notation}

Throughout this work, we use the notation from Chapter 8 of~\cite{Barfoot2017}. A result of this notation that we highlight here includes representing robot pose transformation matrices as an inertial frame (denoted $i$) expressed in the robot body frame (denoted $b$). Given this notation, the kinematic equation for a pose with some generalized body-centric velocity $\boldsymbol{\varpi}_b(t) \in \mathbb{R}^6$ is given by
\begin{align}
    \frac{\partial \boldsymbol{T}_{bi}(t)}{\partial t} &= {\boldsymbol{\varpi}_b(t)}^\wedge \boldsymbol{T}_{bi}(t). 
\end{align} In general, we will not explicitly include the $bi$ and $b$ subscripts. An overview of other notation used in this work can be found in Table~\ref{tab:nomenclature}, with full definitions of relevant operators found in Appendix~\ref{app:notation}.

\begin{table}[!t]
\caption{Nomenclature}\label{tab:nomenclature}
\centering
\begin{tabular}{l p{0.75\columnwidth}}
\hline
\textbf{Symbol} & \textbf{Description} \\
\hline
$SE(3)$ & Special Euclidean group \\
$\mathfrak{se}(3)$ & Lie algebra associated with $SE(3)$ \\
${(\cdot)}^\wedge$ & Linear operator mapping $\mathbb{R}^6$ to $\mathfrak{se}(3)$ \\
${(\cdot)}^\vee$ & Inverse operator of ${(\cdot)}^\wedge$ \\
${(\cdot)}^\curlywedge$ & Spatial cross product operator for $\mathbb{R}^6$ \\
$\boldsymbol{\mathcal{J}}$ & Lie algebra left Jacobian \\
$\boldsymbol{\mathcal{T}}$ & Lie algebra adjoint \\
$\hat{\cdot}$ & Estimated variable \\
$\tilde{\cdot}$ & Measured variable \\
$\check{\cdot}$ & Predicted variable, from prior \\
$\bar{\cdot}$ & Mean value \\
$s$ & Arclength parameter \\
$t$ & Time parameter \\
$\boldsymbol{T}(s,t)$ & Robot pose in $SE(3)$ at $s$ and $t$ \\
$\boldsymbol{\epsilon}(s,t)$ & Body-centric generalized strain vector in $\mathbb{R}^6$ \\
$\boldsymbol{\varpi}(s,t)$ & Body-centric generalized velocity vector in $\mathbb{R}^6$ \\
$\boldsymbol{x}(s,t)$ & Full state vector $\{ \boldsymbol{T}, \boldsymbol{\epsilon}, \boldsymbol{\varpi} \}$ \\
$\boldsymbol{Q}$ & Process noise power spectral density matrix \\
$\boldsymbol{R}$ & Measurement noise covariance matrix \\
$\mathcal{N}(\boldsymbol{\mu}, \boldsymbol{\Sigma})$ & Gaussian distribution with mean $\boldsymbol{\mu}$ and covariance $\boldsymbol{\Sigma}$ \\
$\mathcal{GP}(\boldsymbol{m}, \boldsymbol{k})$ & Gaussian process with mean function $\boldsymbol{m}$ and kernel $\boldsymbol{k}$ \\
\hline
\end{tabular}
\end{table}

\subsection{Setup}

We wish to estimate the continuous-time state of a continuum robot $\boldsymbol{x}(s,t)$ stochastically over both arclength ($s$) and time ($t$). We take a Bayesian inference approach by first defining a prior, then combining measurement data to estimate a posterior distribution for the robot state. We choose a state representation that includes a pose $\boldsymbol{T}(s,t) \in SE(3)$, a body-centric strain $\boldsymbol{\epsilon}(s,t) \in \mathbb{R}^6$, and body-centric velocity of the robot $\boldsymbol{\varpi}(s,t) \in \mathbb{R}^6$. \begin{equation}
    \boldsymbol{x}(s, t) = \{\boldsymbol{T}(s,t), \boldsymbol{\epsilon}(s,t), \boldsymbol{\varpi}(s,t) \}
\end{equation}
    When representing noise and perturbations on the state, $\boldsymbol{x}$, we will often treat $\boldsymbol{x}$ as a vector. While not strictly correct, it will allow later equations to be easier to follow. Specifically,
    \begin{align}
        \boldsymbol{x} &= \bar{\boldsymbol{x}} + \delta \boldsymbol{x}, & \delta \boldsymbol{x} \sim \mathcal{GP}(\boldsymbol{0}, \boldsymbol{P}), 
    \end{align} represents the following per-element operations:
    \begin{align}
        \boldsymbol{T} &= \exp({\delta \boldsymbol{t}}^\wedge) \bar{\boldsymbol{T}}, & \boldsymbol{\epsilon} &= \bar{\boldsymbol{\epsilon}} + \delta \boldsymbol{\epsilon}, & \boldsymbol{\varpi} &= \bar{\boldsymbol{\varpi}} + \delta \boldsymbol{\varpi}
    \end{align}
    with 
    \begin{align}
        \boldsymbol{P} &= \mathbb{E}\left[\begin{bmatrix}
            \delta \boldsymbol{t} \\
            \delta \boldsymbol{\epsilon} \\
            \delta \boldsymbol{\varpi}
        \end{bmatrix}\begin{bmatrix}
            \delta \boldsymbol{t} \\
            \delta \boldsymbol{\epsilon} \\
            \delta \boldsymbol{\varpi}
        \end{bmatrix}^\top\right].
    \end{align}

In a perfect world, we would want to be able to represent the prior for our state as a Gaussian process (GP) in closed-form over time and space, 
\begin{equation}
    \boldsymbol{x}(s, t) \sim \mathcal{GP}(\check{\boldsymbol{x}}(s, t), \check{\boldsymbol{P}}(s,s'; t,t')). 
\end{equation} This work does not propose such a closed-form representation. Rather, we will approximate the prior through the construction and solution of a factor-graph optimization problem, relying heavily on analogy to the time and space systems presented in~\cite{Anderson2015,Lilge2022}. Though this approach may leave the reader longing for a more thorough theoretical justification for the validity of this prior, we will show that such approximations result in reasonable estimates in practice.

The derivation for our estimation framework begins with a continuum robot's kinematics. The basic kinematic expressions for a rod are given by~\cite{Rucker2011, Till2019} \begin{align}
    \frac{\partial \boldsymbol{T}}{\partial t} &= \boldsymbol{\varpi}^\wedge \boldsymbol{T} &
    \frac{\partial \boldsymbol{T}}{\partial s} &= \boldsymbol{\epsilon}^\wedge \boldsymbol{T} &
    \label{eq:consistency}\frac{\partial \boldsymbol{\varpi}}{\partial s} &= \frac{\partial \boldsymbol{\epsilon}}{\partial t} + \boldsymbol{\epsilon}^\curlywedge \boldsymbol{\varpi}
\end{align}
where $\boldsymbol{T} \in SE(3)$ is a transformation matrix, $\boldsymbol{\varpi} \in \mathbb{R}^6$ is the body-centric velocity, and $\boldsymbol{\epsilon} \in \mathbb{R}^6$ is the body-centric strain. Note that each state variable still depends on $s$ and $t$, we drop it here for conciseness. The relation~\eqref{eq:consistency} is a consistency constraint that relates the spatial and temporal derivatives of the body-centric velocity and strain. 

\subsection{Approximate Dynamics}\label{sec:approximate_dynamics}
At the core of our method is a significant simplification of a dynamic Cosserat-rod model made with the assumption that data will be available to compensate for modeling errors. This explicitly sets up our estimation problem as a tradeoff between computation time, prior accuracy, and posterior accuracy. To best illustrate this approximation, consider a dynamic model for a Cosserat rod. We choose to show a simple linear elastic, undamped, gravity-free Cosserat model~\cite{Rucker2011, Till2019}, though a more complex model would demonstrate a similar point. In our notation, these expressions are given by: \begin{align}
    \boldsymbol{K} \frac{\partial \boldsymbol{\epsilon}}{\partial s} + \boldsymbol{\epsilon}^\curlywedge \boldsymbol{K} (\boldsymbol{\epsilon} - \boldsymbol{\epsilon}_0) + \boldsymbol{f} &= \boldsymbol{M} \frac{\partial \boldsymbol{\varpi}}{\partial t} + \boldsymbol{\varpi}^\curlywedge \boldsymbol{M} \boldsymbol{\varpi}.
\end{align} Here, $\boldsymbol{K} \in \mathbb{R}^{6\times 6}$ is a positive-definite stiffness matrix, $\boldsymbol{M} \in \mathbb{R}^{6\times 6}$ is a positive-definite inertial matrix, $\boldsymbol{f} \in \mathbb{R}^{6}$ represents an external wrench, and $\boldsymbol{\epsilon}_0 \in \mathbb{R}^{6}$ is the system strain at rest. 

At this point, the system dynamics are usually solved using a numerical integration method~\cite{Zheng2023,Till2019}. However, these approaches rely on having accurate information not only about the system's physical characteristics but also about the external loads applied. This works well enough in controlled environments but may not be sufficient for operation in more complex and possibly unknown environments. The inherent uncertainty involved in such conditions is what motivates the explicit modeling of uncertainty throughout estimation. We follow the work of~\cite{Anderson2015,Lilge2022}, where we make a seemingly large approximation of the dynamics by injecting white noise onto the second-order derivative terms: 
\begin{subequations}\begin{align}
    \nonumber \frac{\partial \boldsymbol{\varpi}}{\partial t} &= \boldsymbol{M}^{-1} \boldsymbol{K} \frac{\partial \boldsymbol{\epsilon}}{\partial s} + \boldsymbol{M}^{-1} \boldsymbol{\epsilon}^\curlywedge \boldsymbol{K} (\boldsymbol{\epsilon} - \boldsymbol{\epsilon}_0) \\
    \nonumber &\qquad + \boldsymbol{M}^{-1} \boldsymbol{f} - \boldsymbol{M}^{-1} \boldsymbol{\varpi}^\curlywedge \boldsymbol{M} \boldsymbol{\varpi} \\
    &\approx \boldsymbol{w}_1, \\
    \nonumber \frac{\partial \boldsymbol{\epsilon}}{\partial s} &= \boldsymbol{K}^{-1} \boldsymbol{M} \frac{\partial \boldsymbol{\varpi}}{\partial t} + \boldsymbol{K}^{-1} \boldsymbol{\varpi}^\curlywedge \boldsymbol{M} \boldsymbol{\varpi} \\
    \nonumber &\qquad - \boldsymbol{K}^{-1}\boldsymbol{\epsilon}^\curlywedge \boldsymbol{K} (\boldsymbol{\epsilon} - \boldsymbol{\epsilon}_0) - \boldsymbol{K}^{-1}\boldsymbol{f} \\
    &\approx \boldsymbol{w}_2, \\ 
    \intertext{with noise terms defined as}
    \boldsymbol{w}_1(s, t) &\sim \mathcal{GP}(\boldsymbol{0}, \boldsymbol{Q}_1\delta(s - s')\delta(t - t')), \\
    \boldsymbol{w}_2(s, t) &\sim \mathcal{GP}(\boldsymbol{0}, \boldsymbol{Q}_2\delta(s - s')\delta(t - t')).
\end{align}\end{subequations} Here we use a zero-mean Gaussian process white noise with power spectral density matrices $\boldsymbol{Q}_i \in \mathbb{R}^{6 \times 6}$. These are stochastic processes with each point in arclength and time $(s, t)$ being uncorrelated to one another and having a probability density of $\boldsymbol{Q}_i$. Additionally, we inject noise on the cross-derivative terms. This gives us the following approximations:
\begin{subequations}\label{eq:consistency_approximation}\begin{align}
    \frac{\partial \boldsymbol{\epsilon}}{\partial t} &\approx \boldsymbol{w}_3, &
    \frac{\partial \boldsymbol{\varpi}}{\partial s} &\approx \boldsymbol{w}_4,
\end{align}with the noise terms defined as
\begin{align}
    \boldsymbol{w}_3(s, t) &\sim \mathcal{GP}(\boldsymbol{0}, \boldsymbol{Q}_3\delta(s - s')\delta(t - t')), \\
    \boldsymbol{w}_4(s, t) &\sim \mathcal{GP}(\boldsymbol{0}, \boldsymbol{Q}_4\delta(s - s')\delta(t - t')).
\end{align}\end{subequations}
The approximate `dynamics' as presented form the basis for our continuous-time estimation prior. This `white-noise-on-acceleration' (WNOA) approximation results in lower prior information compared to traditional modeling approaches, enabling faster runtime. We will show in this work that such an approximation remains reasonable and that, provided enough sensor information is collected, near state-of-the-art accuracy can be obtained under this model. The underlying ethos of this approach is to trade prior model fidelity for reduced compute, making up for the introduced error through sensor information. This is ultimately justified given the presence of a sensorized robot with unknown dynamics. The tradeoffs of this decision are discussed later in this work.

The reader may realize that~\eqref{eq:consistency_approximation} does not explicitly reflect the structure of the $SE(3)$ consistency constraint of~\eqref{eq:consistency}. We will see later that we will be able to account for this constraint \textit{implicitly} within the structure of the factor graph that will be formed, removing the need to \textit{explicitly} account for it here. 

\subsection{Prior Factors}

We wish to construct a sum of prior factors that when minimized, provides a solution to the approximate dynamics presented in Section~\ref{sec:approximate_dynamics}. We will do this through the use of binary factors (connecting two states) in both time and space. Consider a discrete set of $N$ points in arclength, $s_1, s_2, \ldots, s_N$, and a discrete set of $K$ points in time, $t_1, t_2, \ldots, t_K$. We refer to an arbitrary point in arclength and time using $s_n$ and $t_k$, where $1 \leq n \leq N$ and $1 \leq k \leq K$. These arclengths and times will mark a series of states that we will explicitly estimate and use in the following factors. Afterwards, we will show that we can interpolate between these states in time or space, resulting in a continuous state representation. 

\subsubsection{Time Factors}

We begin by constructing a series of factors that relate the state at a specific point in time, $t_k$, to the state at some later point in time, $t_{k+1}$. Beginning with the time-dynamics of our state variables, 
\begin{align}\label{eq:approx_time_dynamics}
    \frac{\partial \boldsymbol{T}}{\partial t} &= \boldsymbol{\varpi}^\wedge \boldsymbol{T}, &
    \frac{\partial \boldsymbol{\epsilon}}{\partial t} &\approx \boldsymbol{w}_3, &
    \frac{\partial \boldsymbol{\varpi}}{\partial t} &\approx \boldsymbol{w}_1,
\end{align} we can see that these expressions are exactly those of~\cite{Anderson2015} in the time-only GP estimator method. In fact, the pose $\boldsymbol{T}$ and velocity $\boldsymbol{\varpi}$ are directly analogous, with only the addition of the strain $\boldsymbol{\epsilon}$ to the state vector. In the method of~\cite{Anderson2015}, the authors choose to introduce a local pose variable and apply a `white-noise-on-acceleration' prior on the now linear state variables. This works well enough in the time-only and arclength-only cases, but the addition of the consistency constraint in~\eqref{eq:consistency} makes this approach less suitable for our continuous-time estimation framework. Instead, we will construct full nonlinear factors, sacrificing the simplicity of the linearized factors in favor of a more general solution that can be applied to a problem that has arclength and time. Consider an Euler step to the expressions in~\eqref{eq:approx_time_dynamics}, assuming constant strain and velocity at some arbitrary arclength $s_n$, the mean of which is approximately given by
\begin{subequations}\begin{align}
    \bar{\boldsymbol{T}}(s_n, t_{k+1}) &\approx \exp(\Delta t_{k} {\bar{\boldsymbol{\varpi}}(s_n, t_k)}^\wedge)\bar{\boldsymbol{T}}(s_n, t_k), \\
    \bar{\boldsymbol{\epsilon}}(s_n, t_{k+1}) &\approx \bar{\boldsymbol{\epsilon}}(s_n, t_k), \\
    \bar{\boldsymbol{\varpi}}(s_n, t_{k+1}) &\approx \bar{\boldsymbol{\varpi}}(s_n, t_k), \\
    \Delta t_{k} &= t_{k+1} - t_k.
\end{align}\end{subequations}From here we wish to form an error term. A natural choice is to look at the difference between the state at time $t_{k+1}$ and the state at time $t_k$ propagated forward according to the approximate dynamics. This difference reflects a deviation from the approximate dynamics, which we wish to minimize. This error term is given by 
\begin{align}
    \boldsymbol{e}^{(n,k)}_{\text{time}} &= \begin{bmatrix}
        {\ln(\bar{\boldsymbol{T}}_{n,k+1} {\bar{\boldsymbol{T}}_{n,k}}^{-1})}^\vee - \Delta t_{k} \bar{\boldsymbol{\varpi}}_{n,k} \\
        \bar{\boldsymbol{\epsilon}}_{n,k+1} - \bar{\boldsymbol{\epsilon}}_{n,k} \\
        \bar{\boldsymbol{\varpi}}_{n,k+1} - \bar{\boldsymbol{\varpi}}_{n,k} \\
    \end{bmatrix}.
\end{align} With each of these terms, the error is zero when the state progresses between time steps exactly according to the approximate dynamic model. We wish to incorporate a covariance between the state elements through the use of a weighted cost term. In previous works~\cite{Anderson2015,Lilge2022}, such weight matrices are determined through determining the uncertainty added to the system over a specific window of time (or arclength) by integrating the system's differential equations, including a zero-mean GP white noise term. We will use a similar method as these works `by analogy' for our factors, acknowledging that without a closed-form solution to the covariance of our prior, these approaches are not quite equivalent. In practice, however, we demonstrate the ability to tune the presented covariance terms in order to obtain a consistent estimator, pointing towards this approximation being reasonable. For the time factor, the prior covariance accumulation matrix can be derived from the following integration,
\begin{align}
    \boldsymbol{Q}_t(\Delta t_{k}) &= \int_{t_k}^{t_{k+1}} \boldsymbol{\Phi}_t(\Delta \tau) \begin{bmatrix}
        \boldsymbol{0} & \boldsymbol{0} & \boldsymbol{0} \\
        \boldsymbol{0} & \boldsymbol{Q}_3 & \boldsymbol{0} \\
        \boldsymbol{0} & \boldsymbol{0} & \boldsymbol{Q}_1
    \end{bmatrix}{\boldsymbol{\Phi}_t(\Delta \tau)}^\top d\tau \\
    &= \begin{bmatrix}
        \frac{1}{3}{\Delta t}_{k}^3 \boldsymbol{Q}_1 & \boldsymbol{0} & \frac{1}{2}{\Delta t}_{k}^2 \boldsymbol{Q}_1 \\
        \boldsymbol{0} & \Delta t_{k} \boldsymbol{Q}_3 & \boldsymbol{0} \\
        \frac{1}{2}{\Delta t}_{k}^2 \boldsymbol{Q}_1 & \boldsymbol{0} & \Delta t_{k} \boldsymbol{Q}_1
    \end{bmatrix}, \\
    \boldsymbol{\Phi}_t(\Delta t) &= \begin{bmatrix}
        \boldsymbol{I} & \boldsymbol{0} & \Delta t\boldsymbol{I}  \\
        \boldsymbol{0} & \boldsymbol{I} & \boldsymbol{0} \\
        \boldsymbol{0} & \boldsymbol{0} & \boldsymbol{I} 
    \end{bmatrix}.
\end{align} We refer the reader to Chapter 4 of~\cite{Barfoot2017} for a more in-depth tutorial on how this derivation follows from the stochastic differential equations we are adopting for this work. $\boldsymbol{\Phi}_t$ here is the time transition matrix; it will make another appearance when considering interpolation in a later section. Unsurprisingly, this matrix has similarities to that of~\cite{Anderson2015}, with the addition of the new independent strain state. Armed with both the covariance accumulation matrix and a mean error expression, we can form a full weighted cost term, \begin{gather}
    J^{(n,k)}_{\text{time}} = \frac{1}{2}{\boldsymbol{e}^{(n,k)}_{\text{time}}}^\top {\boldsymbol{Q}_t(\Delta t_{k})}^{-1} \boldsymbol{e}^{(n,k)}_{\text{time}}. 
\end{gather}

\subsubsection{Arclength Factors}
We continue by constructing factors for the arclength parameter between states at two points in arclength, $s_n$ and $s_{n+1}$, along the backbone of the robot. Beginning with the arclength dynamics of our state variables, 

\begin{align}\label{eq:approx_space_dynamics}
    \frac{\partial \boldsymbol{T}}{\partial s} &= \boldsymbol{\epsilon}^\wedge \boldsymbol{T}, &
    \frac{\partial \boldsymbol{\epsilon}}{\partial s} &\approx \boldsymbol{w}_2, & 
    \frac{\partial \boldsymbol{\varpi}}{\partial s} &\approx \boldsymbol{w}_4.
\end{align} As before, these expressions are directly analogous to those of~\cite{Lilge2022}. An Euler step is taken to approximate the mean of the state variables along a progression in arclength. The mean of the state variables is approximately given by
\begin{subequations}\begin{align}
    \bar{\boldsymbol{T}}(s_{n+1}, t) &\approx \exp(\Delta s_{n} {\bar{\boldsymbol{\epsilon}}(s_n, t)}^\wedge)\bar{\boldsymbol{T}}(s_n, t) \\
    \bar{\boldsymbol{\epsilon}}(s_{n+1}, t) &\approx \bar{\boldsymbol{\epsilon}}(s_n, t) \\
    \bar{\boldsymbol{\varpi}}(s_{n+1}, t) &\approx \bar{\boldsymbol{\varpi}}(s_n, t), \\
    \Delta s_{n} &= s_{n+1} - s_n.
\end{align}\end{subequations} As before, we can construct the following spatial error expressions for a specific time $t_k$:
\begin{equation}
    \boldsymbol{e}^{(n,k)}_{\text{space}} = \begin{bmatrix}
        {\ln(\bar{\boldsymbol{T}}_{n+1,k} {\bar{\boldsymbol{T}}_{n,k}}^{-1})}^\vee - \Delta s_{n} \bar{\boldsymbol{\epsilon}}_{n,k} \\
        \bar{\boldsymbol{\epsilon}}_{n+1,k} - \bar{\boldsymbol{\epsilon}}_{n,k} \\
        \bar{\boldsymbol{\varpi}}_{n+1,k} - \bar{\boldsymbol{\varpi}}_{n,k}
    \end{bmatrix},
\end{equation} with these terms again equalling zero when the states progress according to the arclength dynamics model. The covariance accumulation matrix for this factor takes on a similar form of the time component, this time relating the strain rate noise uncertainty to the pose over arclength, 
\begin{align}
    \boldsymbol{Q}_s(\Delta s_{n}) &= \int_{s_n}^{s_{n+1}} \boldsymbol{\Phi}_s(\Delta \sigma) \begin{bmatrix}
        \boldsymbol{0} & \boldsymbol{0} & \boldsymbol{0} \\
        \boldsymbol{0} & \boldsymbol{Q}_2 & \boldsymbol{0} \\
        \boldsymbol{0} & \boldsymbol{0} & \boldsymbol{Q}_4
    \end{bmatrix}{\boldsymbol{\Phi}_s(\Delta \sigma)}^\top d\sigma \\
    &= \begin{bmatrix}
        \frac{1}{3}{\Delta s}_{n}^3 \boldsymbol{Q}_2 & \frac{1}{2}{\Delta s}_{n}^2 \boldsymbol{Q}_2 & \boldsymbol{0} \\
        \frac{1}{2}{\Delta s}_{n}^2 \boldsymbol{Q}_2 & \Delta s_{n} \boldsymbol{Q}_2 & \boldsymbol{0} \\
        \boldsymbol{0} & \boldsymbol{0} & \Delta s_{n} \boldsymbol{Q}_4
    \end{bmatrix}, \\
    \boldsymbol{\Phi}_s(\Delta s) &= \begin{bmatrix}
        \boldsymbol{I} & \Delta s \boldsymbol{I} & \boldsymbol{0} \\
        \boldsymbol{0} & \boldsymbol{I} & \boldsymbol{0} \\
        \boldsymbol{0} & \boldsymbol{0} & \boldsymbol{I}
    \end{bmatrix}.
\end{align} The top-left elements of this matrix unsurprisingly echo the result of~\cite{Lilge2022}, with the remaining terms dealing with the new velocity component to the state. We can again form a full weighted cost term, \begin{gather}
    J^{(n,k)}_{\text{space}} = \frac{1}{2}{\boldsymbol{e}^{(n,k)}_{\text{space}}}^\top {\boldsymbol{Q}_s(\Delta s_{n})}^{-1} \boldsymbol{e}^{(n,k)}_{\text{space}}. 
\end{gather}

\subsubsection{Unary Factors}
A third type of factor that we wish to include, which is not novel to this work, but is included for completeness, are unary factors. Unary factors are used to represent the prior knowledge of the system at a specific point in arclength and time. These factors are typically used to represent the initial state, or any other prior knowledge that is known about the system. In the case of continuum robots, these factors can be used to represent information we may have about the base of the robot (e.g., the robot is stationary), the tip (e.g., the strain is zero), or any number of places along the body at any point in time (e.g., the robot is known to pass through a narrow opening). In this work, rather than locking the base state as done in~\cite{Anderson2015, Lilge2022}, we provide a full unary factor at the robot base at each time step as a boundary condition. This approach should more closely reflect the physical reality of a base-mounted CR that may experience small disturbances in its mounting during operation. 

Such a factor assumes that we have some external guess (or just as easily a direct measurement) for the state of the robot $\boldsymbol{x}_0(s, t)$ that represents the actual state corrupted by some noise. Specifically, given an estimate for the state variables at a given arclength $s$ and time $t$, we assume that the estimate is related to the true state via 
\begin{equation}
    \nonumber \boldsymbol{x}_0(s, t) = \begin{bmatrix}
        \boldsymbol{T}_0(s, t) \\
        \boldsymbol{\epsilon}_0(s, t) \\
        \boldsymbol{\varpi}_0(s, t)
    \end{bmatrix} \sim \mathcal{N} \left( \boldsymbol{x}(s, t), \boldsymbol{Q}_{0}(s, t) \right).
\end{equation}
We define an error term for such a factor as\begin{align}
    \boldsymbol{e}^{(n,k)}_{\text{boundary}} &= \begin{bmatrix}
        {\ln(\bar{\boldsymbol{T}}_0(s_n, t_k) {\bar{\boldsymbol{T}}(s_n, t_k)}^{-1})}^\vee \\
        \bar{\boldsymbol{\epsilon}}_0(s_n, t_k) - \bar{\boldsymbol{\epsilon}}(s_n, t_k) \\
        \bar{\boldsymbol{\varpi}}_0(s_n, t_k) - \bar{\boldsymbol{\varpi}}(s_n, t_k)
    \end{bmatrix}.
\end{align} In scenarios where information is only known about part of the state a subset of this error term may be used (e.g.,~the tip strain is known but not the pose or velocity). This term has a cost contribution of 
\begin{gather}
    J^{(n,k)}_{\text{boundary}} = \frac{1}{2}{\boldsymbol{e}^{(n,k)}_{\text{boundary}}}^\top {\boldsymbol{Q}_0(s_n, t_k)}^{-1} \boldsymbol{e}^{(n,k)}_{\text{boundary}}. 
\end{gather}

\subsubsection{Factor Graph}
\begin{figure}[!tp]
    \centering
    \includegraphics[width=\columnwidth]{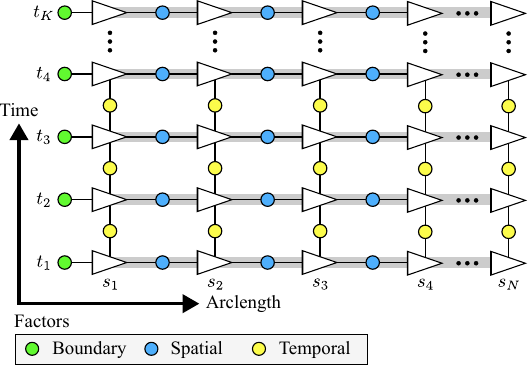}
    \caption{Factor-graph representation of the prior for the continuous-time estimation framework. Unary factors are depicted using green dots, binary space factors are represented with blue dots, and binary time factors are depicted using yellow dots. Each triangle represents a node at a specific point in arclength and time that is estimated directly in our solver. Each factor is connected to one or two nodes, and never connects two nodes that vary in both arclength and time.}\label{fig:factor_graph}
\end{figure}

With each of the three factors introduced, a full factor graph for the prior can be constructed. When thinking about where to assemble specific factors, it is important to remember what the factors are: an Euler approximation for the approximate dynamics of the system. The continuous partial differential equations presented in~\eqref{eq:approx_time_dynamics} and~\eqref{eq:approx_space_dynamics} are theoretically smooth, where, when integrating over arclength and time, the path taken should not matter. Unfortunately, this does not hold true when making Euler approximations. The factor-graph construction should therefore not prefer one integration path over another. A natural choice is to make a grid of nodes across arclength and time, with unary factors used to apply initial or boundary conditions. This approach is taken and depicted in Fig.~\ref{fig:factor_graph}. A grid is assembled using $N$ points in space and $K$ points in time. The nodes of this grid represent states at specific arclengths and times that will be explicitly estimated by the estimator. A boundary condition is applied along the $s = 0$ boundary, corresponding to the physical base of the robot. This results in a factor graph with $NK$ estimation nodes and $N(2K - 1)$ factors. As each node in this graph is only connected to its immediate neighbors, an inherent sparsity is introduced, a fact that is explored further in Section~\ref{sec:solver}.

A subtle assumption being made in this setup is that the consistency constraint,~\eqref{eq:consistency}, will be adequately satisfied without the need for explicitly accounting for it in any loss functions. Neither the temporal factor nor the spatial factor contain any reference to it at all, let alone strictly enforce it. The reason this works is that the structure of the factor graph we form already \textit{implicitly} encodes information from~\eqref{eq:consistency}. The cyclic structure of the graph encourages that a `step in space' followed by a `step in time' results in the same state as a `step in time' followed by a `step in space'. This behavior is ultimately what~\eqref{eq:consistency} encodes, it is the integrability constraint for $SE(3)$. It is important to note that in this framework, \textit{some} amount of flexibility in constraints is \textit{required}. The ethos behind the noise-injection approach proposed in this work is that measurements pull the solution away from one that minimizes the prior error. Strictly enforcing kinematic relationships as constraints adds rigidity to the solver that would harm results. With this in mind, \textit{every} kinematic constraint is expected to have some error to it in the final optimized solution. The tuning of the $\boldsymbol{Q}_i$ and $\boldsymbol{R}_i$ matrices is what influences how constraints are weighted with respect to one another.

\subsection{Measurement Factors}\label{sec:measurement_factors}

We wish to include $M$ noisy measurements from multiple sensors and modalities in this framework. These include direct pose measurements from an electromagnetic pose tracker and rotation rates from a gyroscope. Both of these modalities can be quickly included in our framework, given a measurement model that relates the sensor measurement to the state variables $\boldsymbol{x}(s_n, t_k)$. The following sections will present such models for each of these modalities. 

\subsubsection{Pose Measurements}

We assume that we can take direct pose measurements at a specific arclength down the robot, $s$, at a specific point in time, $t$. Let $\tilde{\boldsymbol{T}}$ represent such a measurement. Following~\cite{Barfoot2014a,Barfoot2017,Lilge2022}, we assume this pose measurement is drawn from a Gaussian with a mean at the true pose, and some zero-mean sensor noise.
\begin{align}
    \tilde{\boldsymbol{T}} &= \exp({\boldsymbol{n}}^\wedge) \boldsymbol{T}(s, t), & \boldsymbol{n} &\sim \mathcal{N}(\boldsymbol{0}, \boldsymbol{R}_{\text{pose}}),
\end{align}
where $\boldsymbol{R}_{\text{pose}}$ is the covariance of the pose sensor noise. Using this setup, we define a measurement error term as the difference between the mean predicted pose and the actual pose measurement, expressed in the Lie algebra as \begin{align}
    \boldsymbol{e}_{\text{pose}} &= {\ln(\tilde{\boldsymbol{T}} {\bar{\boldsymbol{T}}(s, t)}^{-1})}^\vee. 
\end{align}
The total cost associated with a pose measurement is then given by
\begin{equation}
    J_{\text{pose}} = \frac{1}{2}\boldsymbol{e}_{\text{pose}}^\top \boldsymbol{R}_{\text{pose}}^{-1} \boldsymbol{e}_{\text{pose}},
\end{equation}representing the negative log likelihood of the pose measurement. 

\subsubsection{Gyro Measurements}

We assume again that we can take direct measurements of the angular velocity at a specific arclength down the robot, $s$, at a specific point in time, $t$. Let $\tilde{\boldsymbol{\omega}}$ represent such a measurement. We assume this measurement is drawn from a Gaussian with a mean at the true angular velocity, and some zero-mean sensor noise
\begin{align}
    \tilde{\boldsymbol{\omega}} &= \boldsymbol{\omega}(s, t) + \boldsymbol{n}, & \boldsymbol{n} &\sim \mathcal{N}(\boldsymbol{0}, \boldsymbol{R}_{\text{gyro}}),
\end{align}
where $\boldsymbol{R}_{\text{gyro}}$ is the covariance of the gyroscope sensor noise. Noting that the linear velocity, $\boldsymbol{v}$, and angular velocity, $\boldsymbol{\omega}$, are included in our state via $\boldsymbol{\varpi} = \begin{bmatrix}
    \boldsymbol{v} \\
    \boldsymbol{\omega}
\end{bmatrix}$. The error term is then given as 
\begin{equation}
    \boldsymbol{e}_{\text{gyro}} = \tilde{\boldsymbol{\omega}} - \bar{\boldsymbol{\omega}}(s, t),
\end{equation} with a total cost given by
\begin{equation}
    J_{\text{gyro}} = \frac{1}{2}\boldsymbol{e}_{\text{gyro}}^\top \boldsymbol{R}_{\text{gyro}}^{-1} \boldsymbol{e}_{\text{gyro}}.
\end{equation}

While we have presented measurement models for pose and gyroscope sensors, the framework is general enough to include other sensor modalities provided they have a sensor model that relates the measurement to the state variables in $\boldsymbol{x}(s, t)$ and a known Jacobian. FBG sensor~\cite{Lilge2022}, strain gauges~\cite{Ferguson2024}, and stereo cameras~\cite{Anderson2015} have been included in similar frameworks in past works. 

\subsection{Interpolation}

\begin{figure}[!tp]
    \centering
    \includegraphics[width=\columnwidth]{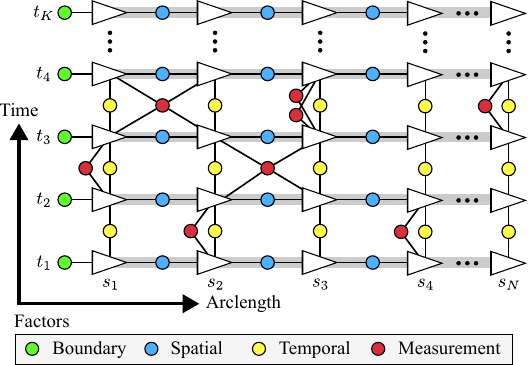}
    \caption{Factor-graph representation of the full continuous-time estimation framework. Unary factors are depicted using green dots, binary space factors with blue dots, and binary time factors with yellow dots. Each triangle represents a node at a specific point in arclength and time that is estimated directly in our solver. Measurements are added sparsely either directly to an estimation node or as an interpolated factor between two estimation nodes and are depicted as red dots. In practice, measurements will solely be added as interpolated factors in time, as the spatial estimation points included in the graph will be selected to align with the physical placement of each sensor on the robot.}\label{fig:factor_graph_measurement}
\end{figure}

In the previous section, the measurement models require access to the state at a specific arclength and time, $\boldsymbol{x}(s, t)$. In practice, we wish to only perform estimation on a discrete set of states ${\{\boldsymbol{x}(s_n, t_k)\}}_{n \in \{1,\ldots,N\},k \in \{1,\ldots,K\}}$. As we wish to include sensors that produce asynchronous measurements with inconsistent timestamps, in addition to being able to represent the robot continuously along its arclength, an interpolation scheme is required.

\subsubsection{Interpolation in Time}

Looking at the time component of our factor graph reveals a similar formulation to that of~\cite{Anderson2015}. This similarity will lend itself to a similar temporal interpolation scheme, one which will be validated in Section~\ref{sec:results_interpolation}. The idea we co-opt from prior works is simple: as the Lie algebra is only valid locally, we cannot have a Gaussian process defined over the entire robot on a single Lie algebra. Rather, we treat each node in the estimator as a reference point in a local Lie algebra, treating the estimate as a string of concatenated Gaussian processes rather than one big one. We then perform GP interpolation on each local Lie algebra. Specifically, we interpolate in time using 
\begin{subequations}\begin{align}
    \boldsymbol{\gamma}_{n,k}(s, t) &= \begin{bmatrix}
        \boldsymbol{\Lambda}_t(s, t) & \boldsymbol{\Psi}_t(s, t) 
    \end{bmatrix} \begin{bmatrix}
        \boldsymbol{\gamma}_{n,k}(s, t_k) \\
        \boldsymbol{\gamma}_{n,k}(s, t_{k+1})
    \end{bmatrix}, \\
    \boldsymbol{\Psi}_t(s, t) &= \boldsymbol{Q}_t(t - t_k) {\boldsymbol{\Phi}_t(t_{k+1} - t)}^\top {\boldsymbol{Q}_t(t_{k+1} - t_k)}^{-1}, \\
    \boldsymbol{\Lambda}_t(s, t) &= \boldsymbol{\Phi}_t(t - t_k) - \boldsymbol{\Psi}_t(s, t) \boldsymbol{\Phi}_t(t_{k+1} - t_k),
\end{align}\end{subequations} where we use a local coordinate, $\boldsymbol{\gamma}_{n,k}$, defined as \begin{subequations}\begin{align}
    \boldsymbol{\gamma}_{n,k}(s, t) &= \begin{bmatrix}
        \boldsymbol{\xi}_{n,k}(s,t) \\
        \frac{\partial}{\partial s}\boldsymbol{\xi}_{n,k}(s,t) \\
        \frac{\partial}{\partial t}\boldsymbol{\xi}_{n,k}(s,t)
    \end{bmatrix}, \\
    \boldsymbol{\xi}_{n,k}(s,t) &= {\ln \left(\boldsymbol{T}(s, t){\boldsymbol{T}(s_n, t_k)}^{-1}\right)}^\vee, \\
    \frac{\partial}{\partial s}\boldsymbol{\xi}_{n,k}(s,t) &= {\boldsymbol{\mathcal{J}}\left(\boldsymbol{\xi}_{n,k}(s,t)\right)}^{-1} \boldsymbol{\epsilon}(s, t), \\
    \frac{\partial}{\partial t}\boldsymbol{\xi}_{n,k}(s,t) &= {\boldsymbol{\mathcal{J}}\left(\boldsymbol{\xi}_{n,k}(s,t)\right)}^{-1} \boldsymbol{\varpi}(s, t),
\end{align} with \begin{align}
    s_n \leq &s \leq s_{n+1}, &
    t_k \leq &t \leq t_{k+1}.
\end{align}\end{subequations} The interpolated state variables can then be expressed as 
\begin{align}
    \nonumber \boldsymbol{x}(s, t) &= \boldsymbol{f}(\boldsymbol{x}_{n,k}, \boldsymbol{x}_{n,k+1}) \\
    &= \begin{Bmatrix}
        \exp({\boldsymbol{\xi}_{n,k}(s,t)}^\wedge) \boldsymbol{T}(s_n, t_k) \\
        \boldsymbol{\mathcal{J}}\left(\boldsymbol{\xi}_{n,k}(s,t)\right) \frac{\partial}{\partial s}\boldsymbol{\xi}_{n,k}(s,t) \\
        \boldsymbol{\mathcal{J}}\left(\boldsymbol{\xi}_{n,k}(s,t)\right) \frac{\partial}{\partial t}\boldsymbol{\xi}_{n,k}(s,t)
    \end{Bmatrix}.
\end{align} 

Covariance interpolation happens after the main solve making use of the Laplace approximation. We can solve for the interpolated covariance with \newpage
\begin{align}
    \nonumber \hat{\boldsymbol{P}}(s, t) &= \check{\boldsymbol{P}}(s, t) + \begin{bmatrix}
        \boldsymbol{\Lambda}_t(s, t) & \boldsymbol{\Psi}_t(s, t) 
    \end{bmatrix}\boldsymbol{P}_{n,k}^{(t)}\begin{bmatrix}
        \boldsymbol{\Lambda}^\top_t(s, t) \\ \boldsymbol{\Psi}^\top_t(s, t) 
    \end{bmatrix}, \\
    \label{eq:cov_square} \boldsymbol{P}_{n,k}^{(t)} &= \begin{bmatrix}
        \boldsymbol{P}_{n,k;n,k} & \boldsymbol{P}_{n,k;n,k+1} \\
        \boldsymbol{P}_{n,k+1;n,k} & \boldsymbol{P}_{n,k+1;n,k+1}
    \end{bmatrix}.
\end{align}

This notably deviates from~\cite{Anderson2015} through the addition of the generalized strain equations, which are introduced using the noise injection approach described in Section~\ref{sec:approximate_dynamics}.

\subsubsection{Interpolation in Space}

In order to perform spatial interpolation, we need only swap out the covariance accumulation and transition matrices from the time case. From here, interpolation along arclength of the estimator is achieved using the same formulation as the time-only interpolation as worked through in the previous section.

\subsubsection{Interpolation in Space-Time}\label{sec:2d_interpolation}

\begin{figure}[!tp]
    \centering
    \includegraphics[width=\columnwidth]{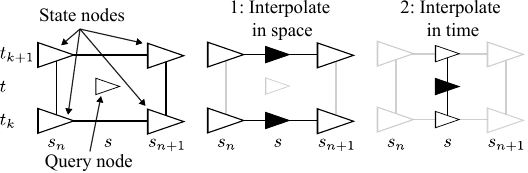}
    \caption{Schematic showing the spatial-temporal interpolation scheme proposed in this work. Given a query node at time $t_k \leq t < t_{k+1}$ and arclength $s_n \leq s < s_{n+1}$, we first use spatial interpolation to find the nodes at $(s, t_{k+1})$ and $(s, t_{k})$. Then, temporal interpolation is applied at each of the previously interpolated nodes, resulting in the estimate at the query point $(s, t)$.}\label{fig:interpolation_schematic}
\end{figure}

In order to interpolate the estimate over both time and space simultaneously, we propose first jointly interpolating two nodes in space, followed by a time-interpolation step. This 2-step process is shown in Fig.~\ref{fig:interpolation_schematic}. We begin by considering some query point $(s, t)$ where $s_n \leq s < s_{n+1}$ and $t_k \leq t < t_{k+1}$. If either equality holds, estimation simply reverts to the methods described in the previous space- or time-only cases. Otherwise, we first perform a joint spatial interpolation to find the states at $(s, t_k)$ and $(s, t_{k+1})$. Afterwards, we use the regular time interpolation to find the state at $(s, t)$. It is important that we perform the spatial interpolation jointly, as doing so independently would not account for the covariance between the interpolated nodes at $(s, t_{k+1})$ and $(s, t_{k})$. The joint interpolation scheme can be written compactly as
\begin{align}
    \begin{bmatrix}
        \boldsymbol{\gamma}_{n,k}(s, t_k) \\
        \boldsymbol{\gamma}_{n,k}(s, t_{k+1})
    \end{bmatrix} &= \boldsymbol{\Lambda}\boldsymbol{\gamma}, \\
    \hat{\boldsymbol{P}}^{(t)}(s, t_k) &= \check{\boldsymbol{P}}^{(t)}(s, t_k) + \boldsymbol{\Lambda} \boldsymbol{P}^{(s,t)}_{n,k} \boldsymbol{\Lambda}^\top,
\end{align} where 
\begin{align}
    \boldsymbol{\Lambda} &= \begin{bmatrix}
        \boldsymbol{\Lambda}_s(s, t_k) & \boldsymbol{\Psi}_s(s, t_k) & \boldsymbol{0} & \boldsymbol{0} \\
        \boldsymbol{0} & \boldsymbol{0} & \boldsymbol{\Lambda}_s(s, t_{k+1}) & \boldsymbol{\Psi}_s(s, t_{k+1})
    \end{bmatrix}, \\
    \boldsymbol{\gamma} &= \begin{bmatrix}
        \boldsymbol{\gamma}_{n,k}(s_n, t_k) \\
        \boldsymbol{\gamma}_{n,k}(s_{n+1}, t_k) \\
        \boldsymbol{\gamma}_{n,k}(s_n, t_{k+1}) \\
        \boldsymbol{\gamma}_{n,k}(s_{n+1}, t_{k+1})
    \end{bmatrix}, 
\end{align} and $\boldsymbol{P}^{(s,t)}_{n,k}$ is the full $72\times 72$ covariance between states at $(s_n, t_k)$, $(s_{n+1}, t_k)$, $(s_n, t_{k+1})$, and $(s_{n+1}, t_{k+1})$. The covariance matrices $\boldsymbol{P}^{(t)}(s, t_k)$ take the form of~\eqref{eq:cov_square} just with respect to arclength $s$ instead of $s_n$. After the joint interpolation is performed, the usual time-only interpolation expression can be applied to its result to achieve full spatio-temporal interpolation. 

\subsubsection{Interpolation of Measurement Factors}\label{sec:interpolated_measurements}

In order to include continuous-time measurements in our factor graph, we must be able to construct Jacobians to the measurement models for the states that are included in the graph. To do this, we make use of the interpolation Jacobians and the chain rule. Specifically, given a measurement model $\boldsymbol{y}(s, t) = \boldsymbol{g}(\boldsymbol{x}(s, t)) + \boldsymbol{n}$ and a time interpolation model $\boldsymbol{x}(s_n, t) = f(\boldsymbol{x}(s_n, t_k), \boldsymbol{x}(s_n, t_{k+1})) + \boldsymbol{w}$, we can construct the Jacobian of the measurement model with respect to the state variables at a specific arclength and times, $\boldsymbol{x}(s_n, t_k), \boldsymbol{x}(s_n, t_{k+1})$, as follows:
\begin{subequations}\begin{align}
    \nonumber \boldsymbol{y}(s, t) &\approx \bar{\boldsymbol{y}}(s, t) + \boldsymbol{n} + \boldsymbol{G}\rvert_{\bar{\boldsymbol{x}}(s, t)} \delta \boldsymbol{x}(s, t) \\
    \nonumber &= \bar{\boldsymbol{y}}(s, t) + \underbrace{\boldsymbol{n} + \boldsymbol{G}\rvert_{\bar{\boldsymbol{x}}(s, t)}\boldsymbol{w}(s, t)}_{\boldsymbol{n}_i} \\
    &\qquad + \boldsymbol{G}\rvert_{\bar{\boldsymbol{x}}(s, t)}\boldsymbol{F}\bigr\rvert_{\bar{\boldsymbol{x}}_{n,k},\bar{\boldsymbol{x}}_{n,k+1}} \begin{bmatrix}
        \delta \boldsymbol{x}_{n,k} \\
        \delta \boldsymbol{x}_{n,k+1}
    \end{bmatrix}, \\
    \boldsymbol{n}_i &\sim \mathcal{N}(\boldsymbol{0}, \boldsymbol{R} + \boldsymbol{G}\rvert_{\bar{\boldsymbol{x}}(s, t)} \boldsymbol{Q}_t {\boldsymbol{G}\rvert_{\bar{\boldsymbol{x}}(s, t)}}^\top),
\end{align}\end{subequations} where $\boldsymbol{G}\rvert_{\bar{\boldsymbol{x}}(s, t)}$ is the measurement model Jacobian evaluated at the interpolated state variables and $\boldsymbol{F}\bigr\rvert_{\bar{\boldsymbol{x}}_{n,k},\bar{\boldsymbol{x}}_{n,k+1}}$ is the interpolation model Jacobian evaluated at the surrounding states. The details for these Jacobians can be found in Appendix~\ref{app:Jacobians}. This formulation allows us to include measurements at any point in time in the estimation problem as long as it is within the estimation window. Fig.~\ref{fig:factor_graph_measurement} shows a factor graph representing such an application of interpolated measurements. It is worth mentioning that this approach implicitly makes an assumption that the interpolated states between two estimation nodes are independent of each other. This has the effect of multiple measurements contributing to overconfidence, with the effect being minimized with smaller choices of $\Delta t$. We find that with most practical choices for $\Delta t$, reasonable estimates are still produced, a result similarly demonstrated in other works making use of the same approximation~\cite{Anderson2015,Lilge2022}. Section~\ref{sec:results_interpolation} explores the ramifications of this assumption in more detail.

\subsection{Solver}\label{sec:solver}

\begin{figure}[!tp]
    \centering
    \includegraphics[width=0.8\columnwidth]{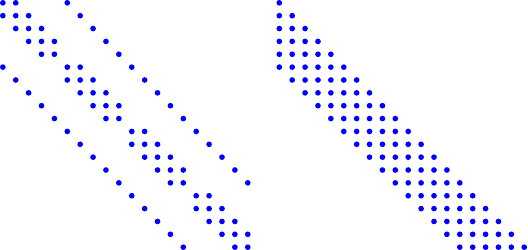}
    \caption{Sparsity structure of the problem's inverse kernel matrix $\check{\boldsymbol{P}}^{-1}$ (left) and its associated Cholesky factor $\check{\boldsymbol{L}}$ (right) for an estimation problem with $N = 5, K = 4$. The sparsity structure follows from the factor graph in Fig.~\ref{fig:factor_graph}. Each row and column of the matrix corresponds to a node in the factor graph, with each non-zero entry corresponding to a correlation between two nodes. The Cholesky factor is lower triangular, with non-zero entries bounded about the diagonal. The band scales with $N$, the number of spatial nodes in our state, leading to a solve complexity of $O(N^3 K)$, where $K$ is the number of time steps in our estimation window. Given $M$ measurements, the total system time complexity, including matrix construction, becomes $O(N^3 K + M)$.}\label{fig:sparsity}
\end{figure}

The total cost function for the factor graph is given by the sum of the prior and measurement costs. Specifically, we have the following cost terms for the prior and measurements:
\begin{subequations}\begin{align}
    J_p &= \sum_k^K \left[J^{(0, k)}_{\text{boundary}} + \sum_n^N \left(J^{(n, k)}_{\text{space}} + J^{(n, k)}_{\text{time}}\right)\right], \\
    J_m &= \sum_m^M J^{(m)}_{\text{pose / gyro}}.
\end{align}\end{subequations} Now we face the challenge of minimizing the cost. Our optimization problem is \begin{subequations}\begin{equation}
    \boldsymbol{x}^* = \arg\min_{\boldsymbol{x}} J_p(\boldsymbol{x}) + J_m(\boldsymbol{x})
\end{equation} where \begin{equation}
    \label{eq:state_stack} \boldsymbol{x} = \begin{bmatrix}
        \boldsymbol{x}_{1,1}^\top &
        \boldsymbol{x}_{2,1}^\top &
        \hdots &
        \boldsymbol{x}_{N,K}^\top
    \end{bmatrix}^\top. 
\end{equation}\end{subequations} The state $\boldsymbol{x}^*$ is then the robot configuration that minimizes the cost function, and is an approximate solution to the dynamics presented in Section~\ref{sec:approximate_dynamics}. The cost is minimized using GTSAM's Levenberg-Marquardt solver to find the solution iteratively, using the factor linearizations presented in Appendix~\ref{app:Jacobians}. The structure of our factor-graph results in an inverse kernel with a block tri-diagonal sparsity pattern shown in Fig.~\ref{fig:sparsity}. This system can be solved efficiently using a Cholesky decomposition, with a total solve complexity of $O(N^3 K)$, where $K$ is the number of time steps in our estimation window. Including the cost of matrix construction, the total time complexity of computing the posterior is $O(N^3K + M)$ where $M$ is the number of measurements included. The proposed solver remains linear in time ($K$), a fact that is important for real-time applications.
As evident from the setup, our estimator is a batch solver. We leave non-batch filter-based extensions as future work, but note that the sparsity structure of the problem should lend itself well to such an approach.

\subsection{Parameter Tuning}\label{sec:param_tuning}

\begin{table}[!t] 
\caption{Estimator Parameters}\label{tab:parameters}
\centering
\begin{tabular}{p{13mm} @{\hspace{2.12mm}} p{13mm} @{\hspace{2.12mm}} p{54.42mm}}
\hline
\textbf{Variable} & \textbf{Size} & \textbf{Description} \\
\hline
$\boldsymbol{Q}_0$ & $18 \times 18$ & Base state covariance \\
$\boldsymbol{Q}_1$ & $6 \times 6$ & Acceleration process noise covariance. Encodes system dynamics \\
$\boldsymbol{Q}_2$ & $6 \times 6$ & Strain rate process noise covariance. Encodes system statics \\
$\boldsymbol{Q}_3$ & $6 \times 6$ & Strain time derivative process noise covariance \\
$\boldsymbol{Q}_4$ & $6 \times 6$ & Velocity spatial derivative process noise covariance \\
$\boldsymbol{R}_{\text{pose}}$ & $6 \times 6$ & Pose measurement noise covariance \\
$\boldsymbol{R}_{\text{gyro}}$ & $3 \times 3$ & Gyroscope measurement noise covariance \\
\hline
\end{tabular}
\end{table}

In the previous sections, several tunable matrices have been introduced. Here, we provide intuition for what effect each has on the final estimator behavior. In general, these matrices encode uncertainty information for the estimator. The more certain (lower value) a number is, the more the estimator will prioritize that information. For the prior terms (which make use of $\boldsymbol{Q}_i$), high confidence relative to the measurement confidences (encoded in $\boldsymbol{R}$) means relying more heavily on the prior. Table~\ref{tab:parameters} presents each tunable matrix, its size, and a brief description. The purpose of noise in this framework is to encode the distribution of effects on the state from unmodeled disturbances. Given that we have selected a simple prior (e.g.,~no physics or actuation inputs), the noise values in this work must be set high to account for the missing information. If additional information is available (e.g.,~tendon forces, stiffness matrix, inertial matrix), it could be incorporated in this framework either through tweaking the prior or as measurement terms to improve the estimate quality. In this work, we opt instead to demonstrate the power of the simplified prior using measurement information directly from robot-mounted sensors, demonstrating competitive results despite the simple model.

In order to tune an estimator, one should start by roughly tuning the prior to reflect the properties that you wish the robot to have. Table~\ref{tab:tuning_heuristics} presents some heuristics for different properties one may wish to achieve. 
\begin{table}[!t]
\caption{Tuning Heuristics}\label{tab:tuning_heuristics}
\centering
\begin{tabular}{p{0.4\columnwidth} @{ $\rightarrow$ } p{0.5\columnwidth}}
\hline
\textbf{Physical Property / Assumption} & \textbf{Parameter Setting} \\
\hline
Kirchhoff rod approximation (minimal shear/elongation) & $\boldsymbol{Q}_0^{7:9}, \boldsymbol{Q}_2^{1:3}$ small, $\boldsymbol{Q}_2^{4:5}$ large \\
Extensible simulated robot & $\boldsymbol{Q}_0^{9}$ large for full-robot extension, $\boldsymbol{Q}_2^{3}$ large for segment-wise extension \\
Fixed base & $\boldsymbol{Q}_0^{1:6}, \boldsymbol{Q}_0^{13:18}$ small \\
Stiff (flexible) robot & $\boldsymbol{Q}_2^{4:5}$ small (large), $\boldsymbol{Q}_4$ small (large) \\
Heavy (light) robot & $\boldsymbol{Q}_1^{1:6}$ small (large), $\boldsymbol{Q}_3$ small (large) \\
Known sensor noise & $\boldsymbol{R} \approx \text{measured noise}$ \\
\hline
\end{tabular}
\end{table}
After these broad characteristics were tuned to be reflected in the estimator, fine-tuning should be performed to adjust where needed. Each system has different information that can inform these parameter values. The robot stiffness, mass, mounting setup, expected actuation profiles, expected contact forces, and downstream applications all can affect the choice of tuning. In general, the ideal tuned parameters are a function of the robot and sensors themselves, and should only have to be found once for each robot.

Two additional parameters that the user has control over are the spatial ($N$) and temporal ($K$) discretization values. Increasing either will reduce the error from the linearizations used throughout this work, leading to convergence in fewer iterations. One should keep in mind that the algorithmic complexity scales with $O(N^3K)$, leading to the practical advice of prioritizing reducing $N$ as low as a given robot allows. In general, the more stiff a robot is, the lower $N$ can be without sacrificing too much accuracy. For most robots, starting with one spatial node every 5--10 cm is a good starting point. Similarly, the parameter $K$ (or more importantly, $K$ per second) should be selected similarly to $N$, looking instead at time evolution of the robot. Light-weight and fast moving robots will result in $K$ per second needing to be increased. For quasi-static robots, using $K$ in the range of 1--5 per second is a good starting point, for more dynamic robots, increase this to 10--30. In general, issues will arise when rotations between two neighboring estimation nodes approach $\frac{\pi}{2}$\,rad, $K$ and $N$ should be large enough to avoid such scenarios.

\section{Apparatus}

\subsection{Simulation}

We use the MuJoCo physics simulator~\cite{Todorov2012} to simulate a two-segment extensible continuum robot. We model the robot as a series of small rigid links, each with a spherical joint and a dimension of extensibility. The robot has a fully contracted length of 30\,cm and can extend to 60\,cm. Six pose sensors and six velocity sensors are equally distributed along the length of the robot and measure noise-free information about the robot for ground truth comparison. For state estimator measurements, two pose sensors, one at the tip and one at the halfway point down the robot length, are corrupted by Gaussian white noise $\boldsymbol{n}$ on $SE(3)$, such that
\begin{align}
    \nonumber \tilde{\boldsymbol{T}}_m &\approx \exp(\boldsymbol{n}_m^\wedge)\tilde{\bar{\boldsymbol{T}}}_m, \\
    \nonumber \boldsymbol{n}_m &\sim \mathcal{N}\left(\boldsymbol{0}, \boldsymbol{R}\right), \\
    \nonumber \boldsymbol{R} &= \text{diag}\{
        5.00 \times 10^{-6}, 5.00 \times 10^{-6}, 5.00 \times 10^{-6}, \\
        &\qquad 1.25 \times 10^{-4}, 1.25 \times 10^{-4}, 1.25 \times 10^{-4}
    \}.
\end{align}

The full MuJoCo XML simulation file is available with the open-sourced code of this project. The data from this simulation is post-processed into three separate scenarios, each of which is described alongside metric performance in Section~\ref{sec:results}. 

\subsection{Prototype}\label{sec:prototype}
We validate our method on a two-segment CR, the details of which are described in this section. The robot is designed to be a testbed for state estimation methods, and sacrifices size and weight for a number of sensors, enabling a ground-truth comparison for the estimator. 

\subsubsection{Robot Design}
The robot prototype is an inverted CR with a hollow nitinol backbone, which limits elongation and contraction. 
Additional PLA 3D-printed modules~\cite{Dewi2024} are used to increase the bending stiffness, with five resin 3D-printed disks used to mount sensors and motion capture markers. The robot is 466\,mm long, 36\,mm in diameter, and consists of two actuated segments. Each segment is actuated using two opposing tendons. 

\subsubsection{Sensors}
\begin{figure}[!tp]
    \centering
    \includegraphics[width=\columnwidth]{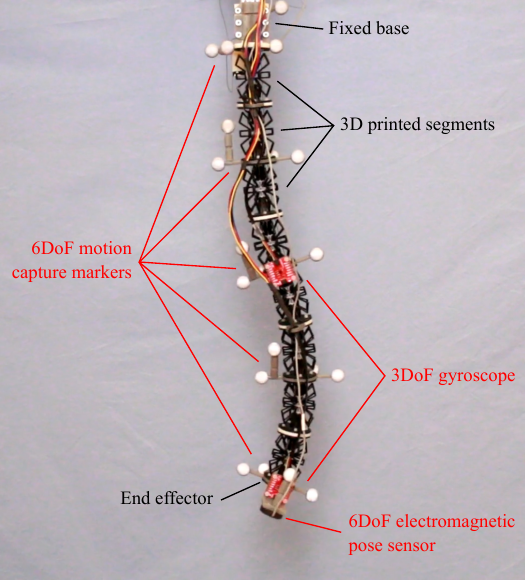}
    \caption{Picture of the robot prototype described in Section~\ref{sec:prototype} during data collection. Sensor locations are labeled in red with other relevant parts labeled in black.}\label{fig:prototype}
\end{figure}

The prototype robot is equipped with nine sensors of three different types. There are five motion capture frames for the Vicon motion capture system (Vicon Motion Systems Ltd., UK) centered on the backbone of the robot, evenly spaced down the length of the robot. This system serves as a ground truth for the robot's pose, providing $\sim$1\,mm position accuracy and $\sim 0.01$\,rad rotational accuracy throughout operation at a rate of 15\,Hz. Two 6-degree-of-freedom (DoF) electromagnetic pose sensors (Aurora v3, Northern Digital Inc., Canada) are mounted on the robot, one at the tip and one at the base. The base sensor is only used for frame alignment and is not fed into the estimator. See the following section for more details. Each collects data at 50\,Hz. Lastly, two 3DoF gyroscopes (ISM330DHCX, STMicroelectronics NV, Netherlands) are mounted at the tip and midpoint of the robot, each collecting data at a rate of 30\,Hz. Fig.~\ref{fig:prototype} shows the robot prototype and sensor placements. 

\subsubsection{Post-Processing}\label{sec:post_processing}

\begin{figure}[!tp]
    \centering
    \includegraphics[width=\columnwidth]{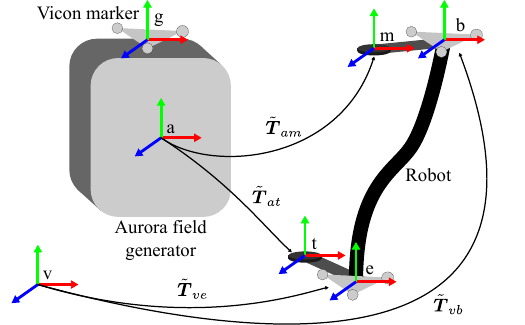}
    \caption{Schematic of relevant coordinate frames for extrinsic calibration. Measured pose values from both the Vicon and Aurora sensing systems are depicted as transforms between two frames. Calibration is performed to estimate the static transforms $\boldsymbol{T}_{\text{mb}}$, $\boldsymbol{T}_{\text{te}}$, and $\boldsymbol{T}_{\text{ag}}$. During estimation, all data is transformed into the Vicon world frame to enable comparison with ground-truth motion capture readings.}\label{fig:frame_definitions}
\end{figure}

Data collected from the sensors is processed after collection and before running the estimator. Outlier rejection and time synchronization of each data type are performed manually. Several static transforms in the system are required for comparing pose estimates to the ground-truth data and are determined through hand-eye calibration. We define the following frames:
\begin{itemize}
    \item v: Vicon world frame
    \item a: Aurora world frame
    \item g: Vicon frame fixed to the Aurora field generator
    \item m: Aurora base marker frame 
    \item t: Aurora tip marker frame 
    \item b: Base frame of the robot
    \item e: End-effector frame of the robot
\end{itemize}
Three unknown static transforms are needed in order to project measured frames into a ground-truth frame: $\boldsymbol{T}_{\text{mb}}, \boldsymbol{T}_{\text{te}}, \boldsymbol{T}_{\text{ag}}$. These transforms represent the extrinsic offsets between frames of interest, and are expected to be static. Fig.~\ref{fig:frame_definitions} depicts the relationship between all frames of interest. We collect pose measurement data for calibration from the robot positioned statically in eight different configurations,
\begin{align}
    \nonumber\boldsymbol{\mathcal{D}} &= \{\tilde{\boldsymbol{T}}_{\text{am}}^{(1)}, \hdots, \tilde{\boldsymbol{T}}_{\text{am}}^{(N_1)}, \tilde{\boldsymbol{T}}_{\text{at}}^{(1)}, \hdots, \tilde{\boldsymbol{T}}_{\text{at}}^{(N_2)}, \\
    &\qquad\tilde{\boldsymbol{T}}_{\text{vb}}^{(1)}, \hdots, \tilde{\boldsymbol{T}}_{\text{vb}}^{(N_3)}, \tilde{\boldsymbol{T}}_{\text{ve}}^{(1)}, \hdots, \tilde{\boldsymbol{T}}_{\text{ve}}^{(N_4)}\}.
\end{align} The unknown transforms $\boldsymbol{T}_{\text{mb}}, \boldsymbol{T}_{\text{te}}$ and $\boldsymbol{T}_{\text{ag}}$ are then found directly using world-hand-eye calibration on this calibration dataset, we choose to use the method from~\cite{Shah2013}. In practice, we see an average reprojection error between these two systems in our setup of $\sim 5$\,mm. This error results in a bias in the ground-truth markers relative to the measurements, contributing to the mean error in the estimates. While bias-estimation remains out-of-scope for this work, we aim to include these biases in the system uncertainties, enabling more consistent results. As this frame alignment is focused at the base and tip of the robot, inconsistencies in the motion capture frames down the robot body are expected to have higher biases.

\subsection{Estimator Parameters}

\begin{figure}[t]
    \centering
    \includegraphics[width=\columnwidth]{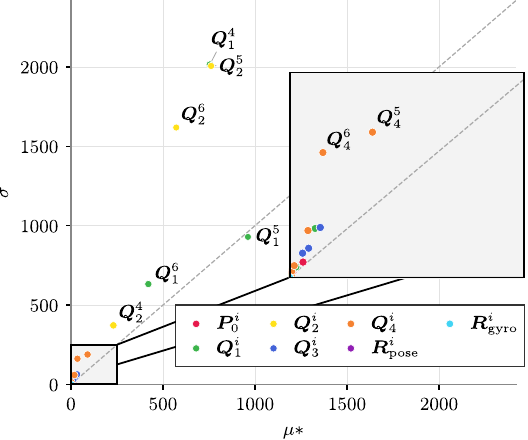}
    \caption{Sensitivity analysis of the NEES metric for the estimator on the `Fast Contact' trajectory using the modified Morris method~\cite{Compolongo2007}. Each point represents that parameter's elementary effect standard deviation ($\sigma$) with respect to the mean absolute elementary effect ($\mu^*$). The sensitivity is found in the information space of each parameter. Elements towards the right indicate stronger influence on consistency, whereas elements towards the top are more nonlinear and dependent on others. In this work, we see that the angular components of $\boldsymbol{Q}_1$ and $\boldsymbol{Q}_2$ have the strongest influence on consistency while also having the most variability in the Morris sense.}\label{fig:sensitivity}
\end{figure}

In order to tune the estimators for both the simulated and experimental robot used in this work, we used the process described in Section~\ref{sec:param_tuning}, fine-tuning on a set of hold-out data. We find that tuning the measurement covariance values slightly noisier than what is observed aids in generating a consistent estimate. This likely comes from the assumption discussed in Section~\ref{sec:interpolated_measurements}, tending to make estimates overconfident, as well as the bias discussed in Section~\ref{sec:post_processing}. Both sets of tuned parameters reflect the robot characteristics of their respective robots, with differences coming from extensibility and inertial properties. All parameters used for both the prototype and the simulated robot estimators in this paper are provided in Appendix~\ref{app:parameters}. In Fig.~\ref{fig:sensitivity}, we perform a sensitivity analysis of the estimator consistency for each of the tunable parameters using the modified Morris method proposed in~\cite{Compolongo2007}. We use 50 trajectories sampled in a parameter space spanning two orders of magnitude below the tuned parameter values (presented in Appendix~\ref{app:parameters}) to two orders of magnitude above the parameter values. We find that the angular components of $\boldsymbol{Q}_1$ and $\boldsymbol{Q}_2$ make up the most prominent contributors to the estimator consistency, a fact that is confirmed qualitatively from the manual tuning process. These are the parameters that are the most sensitive to tuning. A similar trend is observable in the tip error metrics (not shown). While we encourage readers to use the values provided as a starting point for their own estimation problems, fine-tuning on a given robot is required to achieve optimal performance. 

For each experiment trajectory, the estimator discretizes the problem using $N=5$ spatial nodes (or in simulation, $N=7$) and places time nodes at a rate of $30$\,Hz. 
The optimizer parameters are left at their GTSAM default values.
Solve times in this configuration for a 10s trajectory vary between 0.26--0.30\,s on the experimental robot and for a 20\,s simulated trajectory, 0.91--0.92\,s. All experiments are run on a laptop with an Intel i7-13850HX processor and 64\,GB of RAM\@.

The initial state used in the estimation framework for all the results shown here is a straight, stationary robot. The base pose of the robot used in the boundary factors is measured directly. The generalized velocity for the boundary factor is set to zero and the generalized strain to that of a straight robot. Given the mostly convex nature of both the prior and measurement factors proposed here, the configuration used in this work is highly robust to initialization. We see no measurable difference in error metrics between a straight initialization and a random initialization for the data used in this work. This is unlikely to be the case with the addition of measurement models that are non-convex.

\subsection{Benchmarks}
We wish to compare our method to the GP method proposed in~\cite{Lilge2022}. In fact, if we restrict our method to only use a value of $K=1$ and drop the velocity component of the state, our method closely approximates the quasi-static approach. To ensure a fair comparison, we use the same process noise and measurement noise parameters as our method uses in only the spatial dimension. To compare the two methods, only pose measurements are used, as gyro rates cannot be incorporated into a quasi-static method. As the ground-truth measurements are asynchronous to the pose measurements, we only evaluate error metrics using estimates that are within 0.01\,s of a ground-truth reading.

\subsection{Error Metrics}
To evaluate the performance of the estimator, we report three error metrics. As only 6DoF pose information is available as a ground truth, we report errors on position and orientation of the estimator and not on the velocity and strains. Ground-truth data is collected from the Vicon motion capture system asynchronously. During evaluation, the estimator is queried at the same time steps as the ground-truth data to avoid time alignment errors. 

\subsubsection{Estimator Consistency}
First, the consistency of our estimator is evaluated using the average Normalized Estimation Error Squared (NEES) metric~\cite{Barfoot2017}. This metric is used to evaluate the consistency of the covariance estimates produced by the estimator. The average NEES for a given trajectory is computed as follows:
\begin{subequations}\begin{align}
    \boldsymbol{e}_{m} &= {\ln(\tilde{\boldsymbol{T}}_{m}{\boldsymbol{T}_{m}}^{-1})}^\vee, \\
    \label{eq:nees}d_{\text{NEES}} &= \frac{1}{M} \sum_{m=1}^M \boldsymbol{e}_{m}^\top \boldsymbol{P}_{m}^{-1}\boldsymbol{e}_{m}. 
\end{align}\end{subequations} Given we are using 6DoF pose information, a consistent estimator should produce an average NEES value of 6. A value greater than 6 indicates that the estimator is overconfident in its estimates, while a value less than 6 indicates that the estimator is underconfident. In the simulated trajectories, ground-truth velocity data is available, allowing us to define a generalized velocity version of~\eqref{eq:nees}. In this case, an optimal NEES value of 12 reflects a perfectly consistent estimator. The error vector in these cases is simply the difference between the estimated and ground-truth value. Generalized strain errors are not presented as ground-truth values are not easily available due to the pseudo-rigid-body model used in the simulator.

\subsubsection{Accuracy Metrics}
In many practical applications of continuum robots, the tip pose is of primary interest. To remain consistent with the CR literature~\cite{Dong2017, Liu2025, Rone2013}, we report the accuracy of the tip pose using a mean distance metric at the tip of the robot. The position and orientation mean absolute error (MAE) are computed as follows:
\begin{subequations}\begin{align}
    d_{\text{MAE,pos}} &= \frac{1}{M} \sum_{m=1}^M \frac{\lVert\tilde{\boldsymbol{p}}_{m} - {\boldsymbol{p}_{m}}\rVert_2}{L}, \\
    d_{\text{MAE,rot}} &= \frac{1}{M} \sum_{m=1}^M \lVert{\ln(\tilde{\boldsymbol{R}}_{m}{\boldsymbol{R}_{m}^{-1}})}^\vee\rVert_2.
\end{align}\end{subequations}
We similarly present the body pose errors using the same metrics averaged over the four motion capture frames along the length of the robot that coincide with a non-zero length. We also present the root mean square error (RMSE) of the robot tip (and body) pose, which we compute as 
\begin{subequations}\begin{align}
    d_{\text{RMSE,pos}} &= \frac{1}{L} \sqrt{\frac{1}{M} \sum_{m=1}^M \lVert\tilde{\boldsymbol{p}}_{m} - {\boldsymbol{p}_{m}}\rVert^2_2}, \\
    d_{\text{RMSE,rot}} &= \sqrt{\frac{1}{M} \sum_{m=1}^M \lVert{\ln(\tilde{\boldsymbol{R}}_{m}{\boldsymbol{R}_{m}^{-1}})}^\vee\rVert^2_2}.
\end{align}\end{subequations}
Similarly to the NEES metric, we can define an accuracy metric over the generalized velocity, defined as the mean magnitude of error vectors.

\section{Results}\label{sec:results}

\subsection{(Simulation) Robot Extension}
\begin{figure*}[th!]
    \centering
    \includegraphics[width=\textwidth]{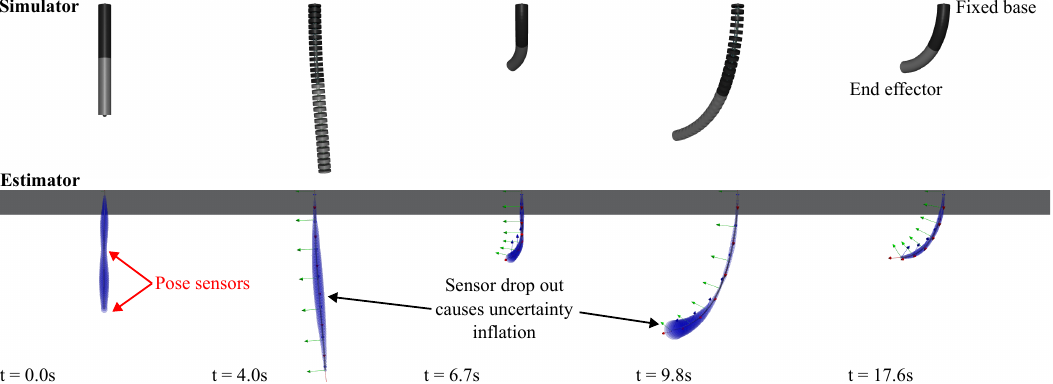}
    \caption{Comparison of the estimator (bottom) and the simulation (top) showing an extensible robot during the `Simulation Dropout' trajectory using two pose sensors at five different time steps. 3$\sigma$ confidence bounds are shown. The estimator demonstrates its ability to generalize to an extensible robot configuration here, highlighting the method's versatility across multiple robot types. The demo shown has an average tip error of 2.0 mm. Time nodes are placed at a rate of 30\,Hz with spatial estimation performed after the fact for visualization.}\label{fig:sim_demo}
\end{figure*}

\begin{table*}[th]
    \centering
    \caption{Performance Metrics and Trajectory Description for Simulated Trajectories.}\label{tab:results_sim}
    \begin{tabular}{p{0.15\textwidth}p{0.03\textwidth}p{0.05\textwidth}p{0.03\textwidth}p{0.05\textwidth}p{0.03\textwidth}p{0.05\textwidth}p{0.03\textwidth}p{0.05\textwidth}p{0.03\textwidth}p{0.2\textwidth}} \hline
        \multicolumn{1}{l}{\multirow{2}{0.1\textwidth}{\textbf{Trial} \\ Data Type}} & \multicolumn{2}{c}{\textbf{MAE (Tip)}} & \multicolumn{2}{c}{\textbf{MAE (Body)}} & \multicolumn{2}{c}{\textbf{RMSE (Tip)}} & \multicolumn{2}{c}{\textbf{RMSE (Body)}} & \multicolumn{1}{c}{\multirow{2}{0.06\textwidth}{\textbf{Average NEES}}} & \multicolumn{1}{c}{\textbf{Trajectory description}} \\
        \multicolumn{1}{c}{} & Pos & Rot & Pos & Rot & Pos & Rot & Pos & Rot & \multicolumn{1}{c}{} & \\ \hline
        \multicolumn{1}{l}{\textbf{Simulation}} & & & & & & & & & 7.44 & \multirow{3}{0.2\textwidth}{A sporadic trajectory with random motions and extensions}\\
        Pose (mm, rad) & 1.555 & 0.005 & 1.205 & 0.021 & 1.935 & 0.005 & 1.773 & 0.029 &  &  \\
        Velocity (mm/s, rad/s) & 28.511 & 0.044 & 15.661 & 0.030 & 45.714 & 0.064 & 30.978 & 0.054 &  & \\ \hline
        \multicolumn{1}{l}{\textbf{Simulation High Noise}} & & & & & & & & & 7.55 & \multirow{3}{0.2\textwidth}{`Simulation' with sensor noise multiplied by a factor of 2}\\
        Pose (mm, rad) & 1.989 & 0.007 & 1.380 & 0.021 & 2.461 & 0.007 & 1.998 & 0.030 &  &  \\
        Velocity (mm/s, rad/s) & 31.583 & 0.048 & 16.702 & 0.032 & 50.530 & 0.068 & 32.381 & 0.055 &  & \\ \hline
        \multicolumn{1}{l}{\textbf{Simulation Dropout}} & & & & & & & & & 7.46 & \multirow{3}{0.2\textwidth}{`Simulation' with the sensors cut off sporadically throughout operation}\\
        Pose (mm, rad) & 2.026 & 0.006 & 1.416 & 0.021 & 2.690 & 0.007 & 2.294 & 0.030 &  &  \\
        Velocity (mm/s, rad/s) & 29.535 & 0.045 & 16.869 & 0.031 & 46.990 & 0.065 & 32.818 & 0.054 &  & \\ \hline
        \end{tabular}
        
        \vspace{0.5em}
        
        \begin{minipage}{\textwidth}
            \raggedright
            \footnotesize
            The NEES metric has an optimal value equal to the number of degrees of freedom in the evaluation. For the simulated results, the optimal value is 12.  
        \end{minipage}
\end{table*}

The estimator is evaluated on each of the extensible simulated scenarios described in previous sections and the results are shown in Table~\ref{tab:results_sim}. The estimate is compared against ground-truth pose and velocity information from the simulator. The estimator is able to produce an estimate of the robot's state, with average tip errors around 2 mm (or 0.7\% of the contracted robot length). A side-by-side comparison of the simulated robot and the estimate result is shown in Fig.~\ref{fig:sim_demo}. The estimator on this trajectory has an average NEES of 7.46, indicating that the covariance estimate is slightly underconfident. The estimator demonstrates a smoothing effect in time when the robot engages in sudden changes in velocity. During this sudden acceleration, the estimator GP smooths the estimate, resulting in a brief period of overconfidence on the estimated velocity. This effect results in the high RMSE errors seen amongst the velocity estimates relative to their MAE errors. Similar smoothing behavior is seen in the past GP estimation works~\cite{Lilge2022} and~\cite{Ferguson2024}.

The high-noise data results in a slightly worse performing estimator than the low noise sensor. The time-smoothing effects of the estimator largely filter out the noise, leaving a smooth estimate as a result. By dropping out the sensor information sporadically, a similar performance is achieved to that of the noisy sensor case. While the estimation accuracy fluctuates between the three trials, the NEES metric remains relatively consistent, showing that the estimator remains consistent in a variety of sensor configurations. 

\subsection{Dynamic Trajectories}
\begin{figure*}[th]
    \centering
    \includegraphics[width=\textwidth]{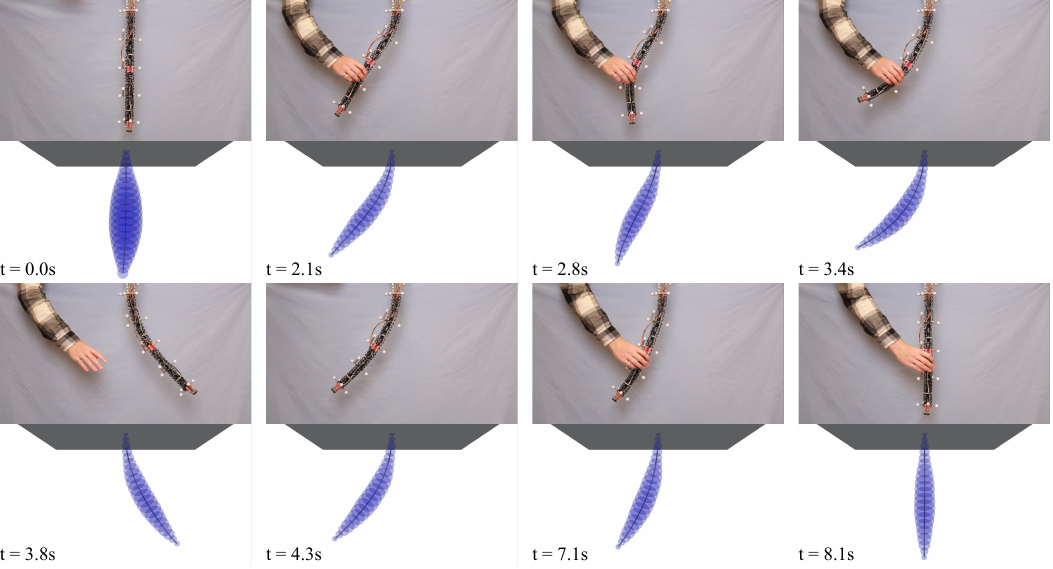}
    \caption{Comparison of the estimator (bottom) and a video (top) recording of the robot during the `Fast Contact' trajectory using both pose and gyroscope sensors at eight different time steps. 3$\sigma$ confidence bounds are shown. The estimator demonstrates its ability to maintain reasonable estimates of the robot's state when unmodeled external forces are applied. The demo shown has an average tip error of 1.169\%. Time nodes are placed at a rate of 30\,Hz with no interpolated states added after the fact.\label{fig:contact_demo}}
\end{figure*}
\begin{figure}[t]
    \centering
    \includegraphics[width=\columnwidth]{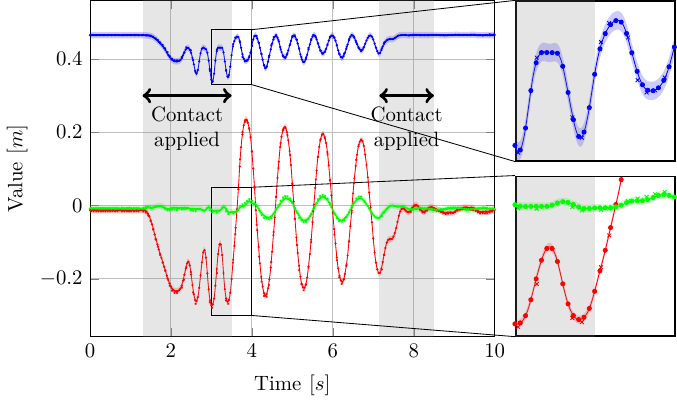}
    \includegraphics[width=\columnwidth]{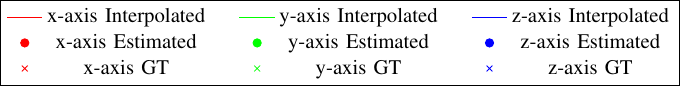}
    \caption{Robot tip position estimate during the `Fast Contact' fused data trajectory. In this demo, external forces are applied to the robot and noted by the shaded regions. Ground-truth values are shown as points, while the estimated mean and 3-$\sigma$ bounds are shown as solid lines and shaded regions, respectively. Interpolation in time is performed after the fact, showing 9 interpolated nodes between each estimated one. The estimator is able to maintain a reasonable estimate of the robot's state despite the unmodeled external forces. We note that during aggressive external forces (bottom zoomed region), the estimator may experience overconfidence in its estimate.}\label{fig:contact_interpolation_demo}
\end{figure}

\begin{table*}[th]
    \centering
    \caption{Performance Metrics and Trajectory Description for Experimental Trajectories.}\label{tab:results_experiment}
    \begin{tabular}{p{0.1\textwidth}p{0.05\textwidth}p{0.04\textwidth}p{0.05\textwidth}p{0.04\textwidth}p{0.05\textwidth}p{0.04\textwidth}p{0.05\textwidth}p{0.04\textwidth}p{0.04\textwidth}p{0.2\textwidth}} \hline
        \multicolumn{1}{l}{\multirow{2}{0.1\textwidth}{\textbf{Trial}\\Configuration}} & \multicolumn{2}{c}{\textbf{MAE (Tip)}} & \multicolumn{2}{c}{\textbf{MAE (Body)}} & \multicolumn{2}{c}{\textbf{RMSE (Tip)}} & \multicolumn{2}{c}{\textbf{RMSE (Body)}} & \multicolumn{1}{c}{\multirow{2}{0.06\textwidth}{\textbf{Average NEES}}} & \multicolumn{1}{c}{\textbf{Trajectory description}} \\
        \multicolumn{1}{c}{} & Pos (\%) & Rot (rad) & Pos (\%) & Rot (rad) & Pos (\%) & Rot (rad) & Pos (\%) & Rot (rad) & \multicolumn{1}{c}{} & \\ \hline
        \multicolumn{1}{l}{\textbf{Out-of-Bounds}} & & & & & & & & & & \multirow{3}{0.2\textwidth}{A dynamic trajectory powered by tendon actuation. Some pose data is lost as the robot leaves the sensor workspace} \\
        Pose-only & 1.565 & 0.043 & 6.566 & 0.132 & 2.318 & 0.059 & 8.290 & 0.177 & \textbf{6.662} & \\
        Gyro-only & 16.735 & 0.745 & 11.712 & 0.158 & 17.939 & 0.762 & 12.988 & 0.201 & 7.778 &  \\
        Fused     & \textbf{1.508} & \textbf{0.038} & \textbf{6.251} & \textbf{0.115} & \textbf{2.091} & \textbf{0.041} & \textbf{7.897} & \textbf{0.151} & 7.933 &  \\ \hline
        \multicolumn{1}{l}{\textbf{Fast Contact}} & & & & & & & & & & \multirow{3}{0.2\textwidth}{A dynamic trajectory achieved through applying sustained external contacts. Fast oscillations present} \\
        Pose-only & 1.170 & \textbf{0.040} & 5.749 & 0.092 & 1.277 & \textbf{0.045} & 7.000 & 0.118 & 4.124 &  \\
        Gyro-only & 16.358 & 0.227 & 13.182 & 0.114 & 19.580 & 0.247 & 15.862 & 0.148 & 3.719 &  \\
        Fused     & \textbf{1.169} & 0.042 & \textbf{5.347} & \textbf{0.072} & \textbf{1.276} & 0.052 & \textbf{6.555} & \textbf{0.090} & \textbf{4.615} &  \\ \hline
        \multicolumn{1}{l}{\textbf{Impulse 1}} & & & & & & & & & & \multirow{3}{0.2\textwidth}{The robot responds to multiple applied impulse contacts. Fast oscillations present} \\
        Pose-only & \textbf{1.562} & 0.043 & 6.266 & 0.107 & \textbf{1.696} & \textbf{0.048} & 6.692 & 0.130 & \textbf{5.455} &  \\
        Gyro-only & 21.866 & 0.276 & 16.710 & 0.224 & 22.453 & 0.278 & 17.918 & 0.269 & 4.835 &  \\
        Fused     & \textbf{1.562} & \textbf{0.042} & \textbf{6.178} & \textbf{0.102} & \textbf{1.696} & \textbf{0.048} & \textbf{6.573} & \textbf{0.120} & 6.769 &  \\ \hline
        \multicolumn{1}{l}{\textbf{Impulse 2}} & & & & & & & & & & \multirow{3}{0.2\textwidth}{The robot responds to multiple applied impulse contacts} \\
        Pose-only & \textbf{1.323} & \textbf{0.059} & 6.786 & 0.119 & \textbf{1.408} & \textbf{0.066} & 8.418 & 0.146 & 6.414 &  \\
        Gyro-only & 21.926 & 0.349 & 16.932 & 0.141 & 24.404 & 0.359 & 19.294 & 0.187 & \textbf{5.752} &  \\
        Fused     & 1.324 & 0.062 & \textbf{6.141} & \textbf{0.088} & 1.409 & 0.072 & \textbf{7.890} & \textbf{0.114} & 7.346 &  \\ \hline
        \multicolumn{1}{l}{\textbf{Slow Free Space}} & & & & & & & & & & \multirow{3}{0.2\textwidth}{A slow trajectory actuated via tendon actuation without any external contact applied} \\
        Pose-only & \textbf{1.102} & \textbf{0.042} & 5.001 & 0.104 & \textbf{1.165} & \textbf{0.042} & 5.344 & 0.138 & 4.809 &  \\
        Gyro-only & 14.550 & 0.834 & 7.122 & 0.113 & 14.650 & 0.843 & 7.438 & 0.141 & 5.044 &  \\
        Fused     & \textbf{1.102} & \textbf{0.042} & \textbf{4.961} & \textbf{0.103} & \textbf{1.165} & \textbf{0.042} & \textbf{5.304} & \textbf{0.136} & \textbf{6.078} &  \\ \hline
        \end{tabular}
        
        \vspace{0.5em}
        
        \begin{minipage}{\textwidth}
            \raggedright
            \footnotesize
            The best performing sensor configuration for each metric on each trial is bolded. For both the MAE and RMSE errors, this is the lowest value. For the NEES metric, this is the value closest to 6, the number of DoFs in the pose states being evaluated. 
        \end{minipage}
\end{table*}

The estimator is evaluated on five different dynamic trajectories, each with varying levels of external contact and operated at different speeds. Table~\ref{tab:results_experiment} contains the performance metrics for each trajectory in addition to a description of each trajectory. In `Fast Contact', large external forces are applied to the robot at two points in the trajectory. A side-by-side comparison of the estimator and the real robot is shown in Fig.~\ref{fig:contact_demo}. The estimator is able to maintain a reasonable estimate of the robot's state despite the unmodeled external forces. The average tip error for this trajectory is 1.169\%, and the average body error is 5.347\%. The estimator has an average NEES of 4.615, indicating that the covariance estimate is slightly underconfident. A closer look at the tip position estimate compared to the ground truth is shown in Fig.~\ref{fig:contact_interpolation_demo}. We note that in the regions where strong external forces are applied, the estimator may experience overconfidence in its estimate. This is a result of the GP smoothing effect, which can cause the estimate to be less responsive to sudden changes in the robot's strains or velocities. When external forces are not applied, the estimator's confidence bounds quickly return to being consistent with the ground truth. Given that no explicit contact model is used, with external forces only being implicitly accounted for in the sensor data, this result is promising and points to this method only improving with additional sensor data (e.g.,~strain sensing). 

Throughout each of the trajectories, the estimator is able to produce reasonable estimates of the robot's state. Average tip errors remain below 1.6\% for all trajectories when pose measurements are used. This is on the order of the noise from the pose sensors used. For most trajectories, we also see that the tip estimate shows no notable improvement with the addition of gyroscope measurements into the system. In the `Out-of-Bounds' trajectory however, where the pose measurements are not available for portions of the trajectory, the gyroscope provides improvement at the tip. In this dropout scenario, the use of both sources of measurements improves the estimate, highlighting the benefit of using multiple sensor modalities. 

\begin{figure}[t]
    \centering
    \includegraphics[width=\columnwidth]{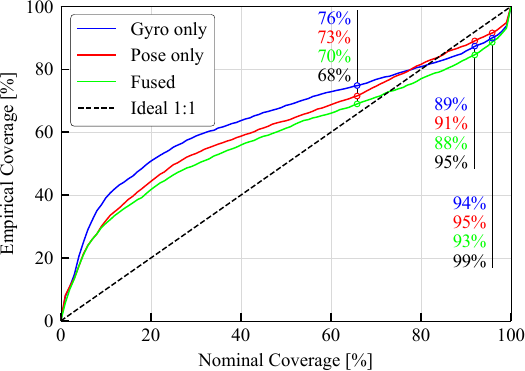}
    \caption{$\chi^2$ coverage plot for the estimator applied over each of the five experimental trajectories. Coverage for the pose-only, gyro-only, and fused configurations are provided. Callouts are provided for the 1-, 2-, and 3-$\sigma$ coverage points. The estimator behavior shows a conservative estimate near the mean with some overconfidence at higher uncertainties, an effect that is exacerbated when the gyroscope sensor is added.}\label{fig:consistency_curves}
\end{figure}

Fig.~\ref{fig:consistency_curves} shows a $\chi^2$ coverage plot evaluated over all the experimental data presented in this work. The plot shows a conservative estimate in all three data scenarios when close to the mean. In outlier regions, a small dip in the graph indicates overconfidence when the mean is far from the ground truth. This result is amplified when using gyroscope data is used in the estimate, a point that is explicitly discussed in the following section.

\begin{figure}[t]
    \centering
    \includegraphics[width=\columnwidth]{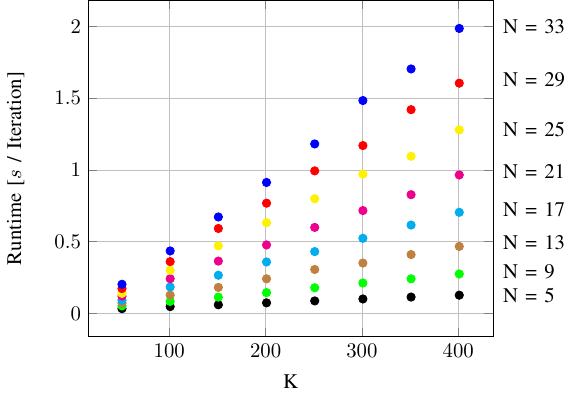}
    \caption{Runtime comparison for the `Fast Contact' fused data estimate. The runtime shown is for estimating only the mean of the state. The proposed implementation runtime scales linearly with respect to the number of time nodes $K$, a result predicted from the sparsity of the problem. Cubic scaling is also observed with respect to the number of spatial nodes $N$, as predicted.}\label{fig:runtime}
\end{figure}

A brief runtime analysis is shown in Fig.~\ref{fig:runtime} for the `Fast Contact' trajectory, estimating only the mean of the state. The implementation scales linearly with respect to the number of time nodes $K$, a result predicted from the sparsity of the problem. Similarly, cubic scaling is observed with respect to the number of spatial nodes $N$, as predicted.

\subsection{Comparison to Quasi-Static GP Prior}

\begin{figure*}[!tp]
    \centering
        \includegraphics[width=\textwidth]{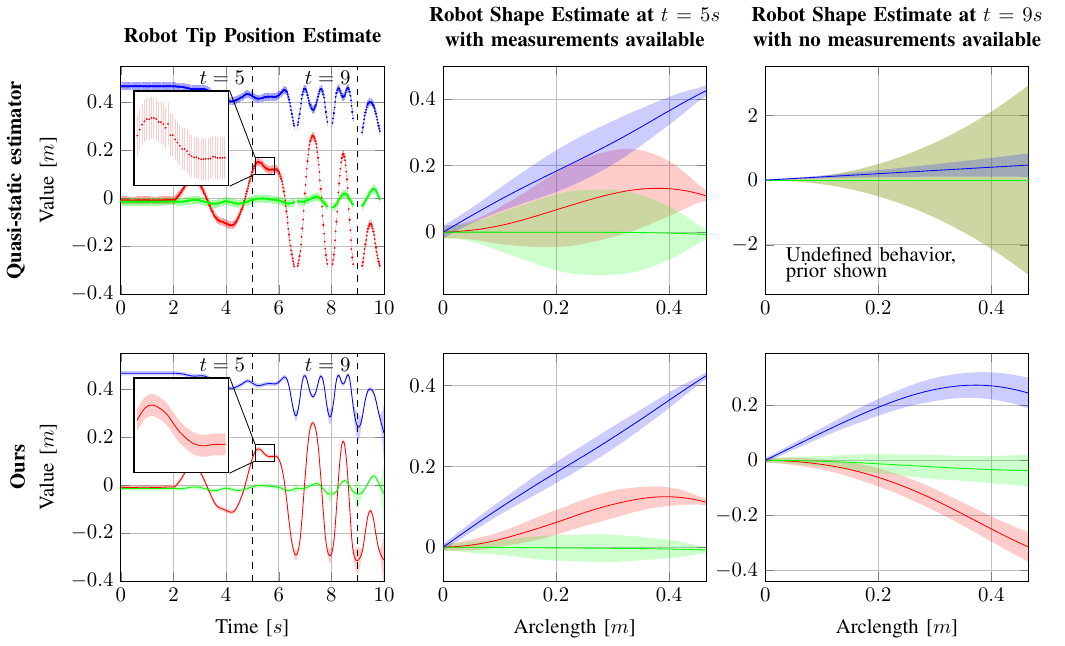}
        \includegraphics{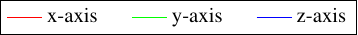}
    \caption{Comparison to the quasi-static benchmark~\cite{Lilge2022} on the `Out-of-Bounds' trajectory using end-effector pose measurements. The leftmost column shows the robot tip position over time for both the proposed method and the quasi-static estimator. The right two columns show a comparison between the two methods at two different time steps, 5\,s and 9\,s. When data at a particular time step is available (e.g.,~middle column), the two methods produce similar mean estimates, with our new method having higher amounts of certainty despite using the same noise characteristics in the arclength WNOA model. The additional information provided in time allows the proposed method to produce a more confident estimate, making it less sensitive to high-frequency noise. When data is not available (e.g.,~right column), the quasi-static benchmark does not have well-defined behavior and is unable to produce a reasonable estimate.}\label{fig:1D_2D_comparison}
\end{figure*}

To compare the two approaches, both are evaluated on the `Out-of-Bounds' trajectory. The quasi-static method is applied to each time step containing a measurement with $N = 5$ spatial nodes. Fig.~\ref{fig:1D_2D_comparison} contains the results of both methods. As expected, the quasi-static method is able to produce reasonable estimates, but is sensitive to high-frequency noise in the sensor data over time. This is clear when zooming in on the time series data for the tip position, where visible jitter is present. In contrast, the proposed method is able to smooth out the estimate in time. When evaluated on the set of ground-truth data within 0.01s of the quasi-static estimates, the quasi-static estimator achieves a tip-position RMSE accuracy of 2.05\% while the proposed spatio-temporal estimator achieves 1.26\% on the same set of data. Similarly, tip-rotation RMSE is 0.044\,rad for the quasi-static estimator and 0.035\,rad for the spatio-temporal one. Another region of interest is when no data is available. In `Out-of-Bounds' this occurs when the robot tip leaves the Aurora workspace. During this time, the quasi-static method does not have a well-defined behavior. The proposed method, however, is able to generate a reasonable estimate while accounting for the increased uncertainty in its confidence bounds. 

More interestingly, the proposed method also produces a more confident estimate even when data is available. This comes from the fact that it can leverage information at other points in time to refine its estimate. This is particularly evident in the case shown, which is solved using batch estimation, but a similar effect is expected in a sliding window implementation, albeit while lacking information about future data points.

\subsection{Interpolation Validation}\label{sec:results_interpolation}
\begin{figure}[!tp]
    \centering
    \includegraphics[width=\columnwidth]{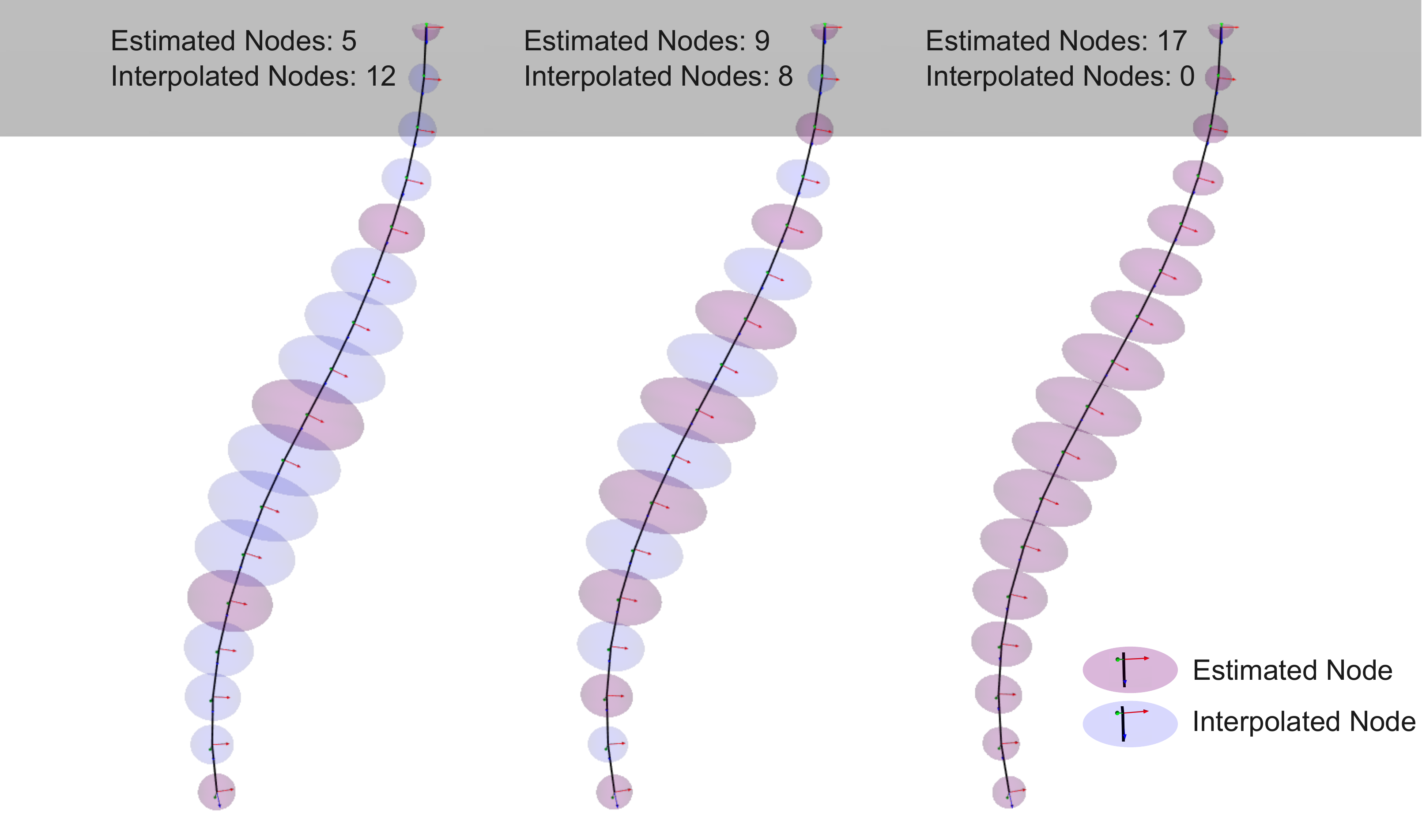}
    \caption{A snapshot of the `Fast Contact' fused data estimate at time 2.833\,s with three different selections of $N$. For each of the $N = 5, 9, 17$ cases, interpolation is performed so that $17$ total nodes are displayed. As $N$ increases, the covariance estimates slightly decrease. The change also results in small deviations in the mean estimate.}\label{fig:interpolation}
\end{figure}
While interpolation in time is performed constantly during the solve in order to include asynchronous measurements, the states are not explicitly stored. After the solve, however, we can interpolate in time and space. Fig.~\ref{fig:contact_interpolation_demo} shows the time-interpolated tip position estimate during the `Fast Contact' trajectory. Each estimated node contains 9 interpolated nodes in between, resulting in the display of 300 nodes / second. Because of the `white-noise-on-acceleration' assumption in the models from Section~\ref{sec:approximate_dynamics}, the interpolated states form a series of cubic splines in the associated Lie algebra, which remain smooth in time. 

Similarly, Fig.~\ref{fig:interpolation} shows interpolation in space for the same trajectory at a single time step. A similar cubic spline behavior is observed with the interpolated states. When solving the same problem with different numbers of spatial nodes $N$, we see that the covariance estimate decreases slightly as $N$ increases, and the mean estimate changes slightly. Some change is expected as with more spatial nodes, the Euler step approximation used in the creation of the prior factors takes a smaller step, resulting in a more finely discretized system. We can quantify this effect to get an approximation for the order of variation introduced by a choice of discretization by treating one estimate using fully estimated nodes as a `ground truth' and comparing solves that replace estimated nodes with interpolated ones. For the cases in Fig.~\ref{fig:interpolation}, we do just this and find that the $N=5$ case has an average `tip error' of 0.11\% and a `body error' of 0.38\% while the $N=9$ case has an average `tip error' of 0.06\% and a `body error' of 0.21\%. While this result does not provide insight into the accuracy effects of the choice of $N$, it does indicate that the larger the discretization step used, the more deviation from a more finely discretized solution we can expect. Users of the method should keep this in mind when selecting a value of $N$ for their application.

\section{Discussion}
This work introduces continuous-time stochastic estimation to the domain of continuum robots. By providing an uncertainty-aware estimate, one can better make decisions surrounding required sensor placements and collection rates given a specified task. We hope by introducing uncertainty information into the continuum robotics field, planning and controls work will be able to integrate this information into future approaches, leading to safer robot operation. High uncertainty could be used to enact more conservative controls or adjust trajectory planners to achieve improved confidence. While we chose to use a configuration tuned towards as close to an NEES value of six on our hold-out set as possible, this came at the expense of being overconfident in some higher contact scenarios. The hold-out itself does not have high-contact or sensor-dropout effects. Users of this framework may wish to err on the side of underconfidence if such high-impact scenarios are expected in their application. 

In our experimental setup we have a high-rate pose sensor and a gyroscope collocated at the tip. Because the pose sensor is high-rate, angular velocity information is already implicitly encoded, leading to the tip estimate performance not being improved when adding the gyroscope data. The body frame errors tell a different story, where adding gyroscope data at the midpoint consistently improves the body estimates. This corresponds to the initial claim in this paper that in order to use this simplified prior, additional sensor information is required to fill in the lost information. Interestingly, Fig.~\ref{fig:consistency_curves} indicates that the addition of gyroscope information results in overconfidence at the distribution tails. Our best guess as to where this comes from is from an unaccounted for bias in extrinsic transforms between the supposedly collocated pose sensor and gyroscope at the tip. This type of overconfidence at the tail regions is common for estimation algorithms that do not perfectly model the system they estimate (such as when an extrinsic offset is askew). As uncertainty is ultimately an estimate of the distribution of states possible because of unmodeled effects on the system, users of this framework should tune the $\boldsymbol{Q}$ matrices to reflect the types of disturbances expected in their particular use-case. The closer to in-distribution a hold-out set is, the better.

We qualitatively validate the theoretical runtime properties of the proposed estimator, demonstrating linear scaling with respect to the number of time nodes and cubic scaling with respect to the number of spatial nodes. Also demonstrated is competitive accuracy ($<2$\% tip error) with good runtime performance, with each 10s trial consistently being solved in well under 1s. While convergence will vary based on convergence thresholds, robot speeds, and discretization, for the robot in this work, this results in a per time step solve time of $<2$ms. This result is demonstrated across a variety of robot motions, external contacts, and two different robot types (extensible and non-extensible). We demonstrate the inclusion of multiple asynchronous sensor modalities and open the possibility for the addition of any number of other sensors, highlighting the flexibility of the framework. Additional sensor modalities remain simple to add to the proposed framework, only requiring a measurement error model that can be linearized. In most trials, fusing gyroscope data does not provide much additional information to the pose sensor only case, except in the `Out-of-Bounds' trajectory. For this trajectory, including additional sensor modalities provides an element of robustness to partial sensor failure, showing the gyroscope data can help fill in the gaps left by the pose-sensor dropout.

\subsection{Limitations and Future Work}

The approximations made in this work follow similar assumptions to those made in previous continuous estimation works~\cite{Anderson2015,Lilge2022}. These include the approximate linearizations of relevant Jacobians and the independence assumption between the interpolated states used in the measurement models. The latter of these assumptions does have a notable impact on the state estimate, a fact that becomes prohibitively worse with lower estimation node rates. As with other factor-graph methods, the method proposed here could be converted into an online filter through the iterative extension and marginalization of the graph. The $<1$s performance suggests that any such filter would be capable of running in real-time, an extension we leave for future work. Apart from the motivating works, we make further assumptions surrounding the dynamic Cosserat rod model via the WNOA Cosserat rod dynamics. This choice surely limits the accuracy of the estimator compared to the more sophisticated models of~\cite{Rucker2011,Zheng2023}, a trade-off that may be worth it given the flexibility of the estimator when handling asynchronous input data streams and its additional covariance estimation capacity. The estimator is expected to fail when the external disturbances grow larger than what is tuned for in the $\boldsymbol{Q}$ matrices. Early signs of this can be seen in the overconfidence of the `Impulse 2' trajectory, where significant external forces (relative to the hold-out set) are applied.

Due to the nature of the sensors used in this work, the results shown are relatively robust to initialization conditions. When initializing the robot to be straight and stationary, the batch estimator with the given tuning always converges. The addition of sensors whose measurement models may introduce singularities or local minima will require additional thought when initializing. 

As the pose factors employed in this work use the Lie algebra $\mathfrak{se}(3)$ as the error space, extremely large deviations in angular components between states can cause issues. This is easily mitigated through increasing the spatial discretization $N$ if too much bending occurs or in $K$ if the robot bends too quickly. Extremely long and flexible robots (e.g.,~everting vine robots) are most at risk of hitting computational bottlenecks for any future online applications, as they would require a high number of spatial discretizations, scaling the algorithmic complexity with $O(N^3)$.  

While the state included in this work only consists of pose, generalized strain, and generalized velocity, the factor-graph formulation easily allows for the addition of other states (e.g.,~biases, extrinsic and intrinsic parameters, temperature, other robots). We encourage future work to exploit this fact to rapidly expand the capabilities of the presented framework.   

\section{Conclusion}

In this work, we presented a fully stochastic, continuous-time estimation framework for continuum robots by extending the works of~\cite{Lilge2022} and~\cite{Anderson2015} to the domain of continuum robot trajectory estimation (i.e., space and time). We demonstrated the ability of the framework to fuse asynchronous measurements from multiple sensor modalities in an uncertainty-aware way. Alongside this work, we provide an open-source implementation of our method to the research community at \url{https://github.com/utiasASRL/space_time_continuum}.

By formulating the continuous-time trajectory estimation of a continuum robot as a factor-graph optimization problem, we hope to open the door to further cross-domain adaptation from works established in mobile robotics. This framework provides a flexible foundation for the addition of other sensors such as cameras, IMUs, or strain gauges, as well as the incorporation of more complex factor-graph-based algorithms used for localization, mapping, and sliding window filtering. To the best of our knowledge, this is the first stochastic continuous-time state estimation framework for continuum robots, and we hope it will serve as a useful tool for the community to build upon.

\begin{appendices}
  \section{Notation}\label{app:notation}
Throughout this work we use the notation of~\cite{Barfoot2017}. Mean values are represented with a bar, $\bar{x}$. Posterior values are represented with a hat, $\hat{x}$. Lastly, prior values are represented with a check, $\check{x}$. Lower case, non-bolded (e.g.,~$x$) characters represent scalars, vectors are lower case, bolded (e.g.,~$\boldsymbol{x}$), and matrices are upper case, bolded (e.g.,~$\boldsymbol{X}$). We define the hat and curly hat operators for the three and six dimensional vectors, $\boldsymbol{x}, \boldsymbol{\rho}, \boldsymbol{\phi} \in \mathbb{R}^3, \boldsymbol{\xi} = \begin{bmatrix}
        \boldsymbol{\rho}^\top &
        \boldsymbol{\phi}^\top
    \end{bmatrix}^\top \in \mathbb{R}^6 $, as
\begin{align}
    \boldsymbol{x}^\wedge &= \begin{bmatrix}
        0 & -x_3 & x_2 \\
        x_3 & 0 & -x_1 \\
        -x_2 & x_1 & 0
    \end{bmatrix},
\end{align}
\begin{align}
    \boldsymbol{\xi}^\wedge &= \begin{bmatrix}
        \boldsymbol{\phi}^\wedge & \boldsymbol{\rho} \\
        \boldsymbol{0}^\top & 0 
    \end{bmatrix}, &
    \boldsymbol{\xi}^\curlywedge &= \begin{bmatrix}
        \boldsymbol{\phi}^\wedge & \boldsymbol{\rho}^\wedge \\
        \boldsymbol{0} & \boldsymbol{\phi}^\wedge
    \end{bmatrix}.
\end{align}The inverse of the wedge operator is defined as
\begin{align}
    {\left(\boldsymbol{X}^\vee\right)}^\wedge &= \boldsymbol{X}, & {\left(\boldsymbol{x}^\wedge\right)}^\vee &= \boldsymbol{x}.\label{eq:vee}
\end{align}

All transformations are represented as $\boldsymbol{T}_{bi} \in SE(3)$, denoting the inertial frame pose relative to the body frame. A natural consequence of this notation that differs from many works in the continuum robotics literature is that the Jacobian $\boldsymbol{\mathcal{J}}$ is the left Jacobian of $SE(3)$ and that body-frame kinematics are applied to the left side of a transformation. For example, the kinematic equation for a pose with some body-centric velocity $\boldsymbol{\varpi}_b(t) \in \mathbb{R}^6$ is given by
\begin{align}
    \frac{\partial \boldsymbol{T}_{bi}(t)}{\partial t} &= {\boldsymbol{\varpi}_b(t)}^\wedge \boldsymbol{T}_{bi}(t). 
\end{align} In general, we will not explicitly include the $bi$ and $b$ subscripts. A superscript containing a number inside of parentheses, (e.g.,~$\boldsymbol{x}^{(i)}$) denotes the $i$th index of that object and will be used when conflicting super/subscripts are present.
  \section{Jacobians}\label{app:Jacobians}
\newcommand{\jacinv}{\boldsymbol{\mathcal{J}}_{n,k}^{-1}}
\newcommand{\ad}{\boldsymbol{\mathcal{T}}_{n,k}}
\newcommand{\jactinv}{{\boldsymbol{\mathcal{J}}_{n,k}^{(t)}}^{-1}}
\newcommand{\adt}{\boldsymbol{\mathcal{T}}_{n,k}^{(t)}}

\subsection*{Binary Factor}
The binary time factor error term is linearized as
\begin{subequations}
\begin{align}
    \boldsymbol{e}^{(n,k)}_{\text{time}} &\approx \bar{\boldsymbol{e}}^{(n,k)}_{\text{time}} +\begin{bmatrix}
         \boldsymbol{F}^{(n,k)}_{\text{time}} & 
         \boldsymbol{E}^{(n,k)}_{\text{time}}
    \end{bmatrix} \begin{bmatrix}
        \delta \boldsymbol{x}_{n, k} \\
        \delta \boldsymbol{x}_{n, k+1}
    \end{bmatrix}, \\
    \boldsymbol{F}^{(n,k)}_{\text{time}} &= \begin{bmatrix}
        -\jactinv \adt & \boldsymbol{0} & - \Delta t \boldsymbol{I} \\
        \boldsymbol{0} & -\boldsymbol{I} & \boldsymbol{0} \\
        \boldsymbol{0} & \boldsymbol{0} & -\boldsymbol{I}
    \end{bmatrix}, \\
    \boldsymbol{E}^{(n,k)}_{\text{time}} &= \begin{bmatrix}
        \jactinv & \boldsymbol{0} & \boldsymbol{0} \\
        \boldsymbol{0} & \boldsymbol{I} & \boldsymbol{0} \\
        \boldsymbol{0} & \boldsymbol{0} & \boldsymbol{I}
    \end{bmatrix}, \\
    \boldsymbol{\mathcal{J}}^{(t)}_{n,k} &= \boldsymbol{\mathcal{J}}({\ln(\boldsymbol{T}_{n,k+1}\boldsymbol{T}_{n,k}^{-1})}^\vee), \\
    \boldsymbol{\mathcal{T}}^{(t)}_{n,k} &= \text{Ad}(\boldsymbol{T}_{n,k+1}\boldsymbol{T}_{n,k}^{-1}).
\end{align}
\end{subequations}

The space factor term is linearized in the same way with strain as the active derivative term in place of velocity.

\subsection*{Unary Factor} 
The unary error terms are linearized as
\begin{subequations}
\begin{align}
    \boldsymbol{e}^{(n,k)}_{\text{boundary}} &\approx \bar{\boldsymbol{e}}^{(n,k)}_{\text{boundary}} + \boldsymbol{E}^{(n,k)}_{\text{boundary}} \delta \boldsymbol{x}_{n, k}, \\
    \boldsymbol{E}^{(n,k)}_{\text{boundary}} &= \begin{bmatrix}
        -\jacinv \ad & \boldsymbol{0} & \boldsymbol{0} \\
        \boldsymbol{0} & -\boldsymbol{I} & \boldsymbol{0} \\
        \boldsymbol{0} & \boldsymbol{0} & -\boldsymbol{I}
    \end{bmatrix}, \\
    \boldsymbol{\mathcal{J}}_{n,k} &= \boldsymbol{\mathcal{J}}({\ln(\boldsymbol{T}_0^{(n,k)}\boldsymbol{T}_{n,k}^{-1})}^\vee), \\
    \boldsymbol{\mathcal{T}}_{n,k} &= \text{Ad}(\boldsymbol{T}_0^{(n,k)}\boldsymbol{T}_{n,k}^{-1}).
\end{align}
\end{subequations}

\subsection*{Measurement Factors}
The pose measurement error terms are linearized as
\begin{subequations}
\begin{align}
    \boldsymbol{e}_{\text{pose}} &\approx \bar{\boldsymbol{e}}_{\text{pose}} + \boldsymbol{E}_{\text{pose}} \delta \boldsymbol{x}_{n, k} \\
    \boldsymbol{E}_{\text{pose}} &= \begin{bmatrix}
        -\tilde{\boldsymbol{\mathcal{J}}}_{n,k} \tilde{\boldsymbol{\mathcal{T}}}_{n,k} & \boldsymbol{0} & \boldsymbol{0}
    \end{bmatrix}, \\
    \tilde{\boldsymbol{\mathcal{J}}}_{n,k} &= \boldsymbol{\mathcal{J}}({\ln(\tilde{\boldsymbol{T}}_{n,k}\boldsymbol{T}_{n,k}^{-1})}^\vee), \\
    \tilde{\boldsymbol{\mathcal{T}}}_{n,k} &= \text{Ad}(\tilde{\boldsymbol{T}}_{n,k}\boldsymbol{T}_{n,k}^{-1}).
\end{align}
\end{subequations}

The gyro measurement error terms are linearized as
\begin{subequations}
\begin{align}
    \boldsymbol{e}_{\text{gyro}} &\approx \bar{\boldsymbol{e}}_{\text{gyro}} + \boldsymbol{E}_{\text{gyro}} \delta \boldsymbol{x}_{n, k}, \\
    \boldsymbol{E}_{\text{gyro}} &= \begin{bmatrix}
        \boldsymbol{0} & \boldsymbol{0} & \begin{bmatrix}
            \boldsymbol{0} & \boldsymbol{0} \\ 
            \boldsymbol{0} & \boldsymbol{I}
        \end{bmatrix}
    \end{bmatrix}.
\end{align}
\end{subequations}

\subsection*{Interpolation Jacobians}
The following Jacobians are the approximate linearized Jacobians for the time-interpolation. Spatial interpolation Jacobians will take a similar form. Throughout this section, we use the box symbol, ${}^\square$, to indicate a stacked vector of two boundary nodes in a given interpolation. We wish to represent a perturbation to an interpolated state with respect to a perturbation of the estimation states. That is, 
\begin{subequations}
\begin{align}
    \delta \boldsymbol{x}(s, t) &\approx \frac{\partial \boldsymbol{x}(s, t)}{\partial \boldsymbol{x}_{n,k}^\square} \delta \boldsymbol{x}_{n,k}^\square, \\
    \boldsymbol{x}^\square_{n,k} &= \begin{bmatrix}
        \boldsymbol{x}_{n,k} \\
        \boldsymbol{x}_{n,k+1}
    \end{bmatrix}.
\end{align}
\end{subequations}
The Jacobians are
\begin{subequations}
\begin{align}
    \delta \boldsymbol{t}_{s, t} &\approx \underbrace{\left( { \boldsymbol{\mathcal{J}}}_{s,t}\boldsymbol{P}_1 \boldsymbol{\Lambda}_{s, t}\frac{\partial \boldsymbol{\gamma}_{n,k}^\square}{\partial \boldsymbol{x}_{n,k}^\square}  + \boldsymbol{\mathcal{T}}_{s, t} \boldsymbol{P}^*_1 \right)}_{\frac{\partial \boldsymbol{t}_{s, t}}{\partial \boldsymbol{x}_{n,k}^\square}}\delta \boldsymbol{x}_{n,k}^\square, \\
    \delta \boldsymbol{\varpi}_{s, t} &\approx \underbrace{(\boldsymbol{\mathcal{J}}_{s, t} \boldsymbol{P}_3 - \frac{1}{2}{(\boldsymbol{P}_3\boldsymbol{\Lambda}_{s, t}\bar{\boldsymbol{\gamma}}^\square_{n,k})}^\curlywedge \boldsymbol{P}_1)\boldsymbol{\Lambda}_{s, t}\frac{\partial \boldsymbol{\gamma}_{n,k}^\square}{\partial \boldsymbol{x}_{n,k}^\square}}_{\frac{\partial \boldsymbol{\varpi}_{s, t}}{\partial \boldsymbol{x}_{n,k}^\square}} \delta \boldsymbol{x}_{n,k}^\square,
\end{align} and $\delta \boldsymbol{\epsilon}_{s, t}$ takes the form of $\delta \boldsymbol{\varpi}_{s, t}$ with the relevant velocity states swapped for strain. $\boldsymbol{P}_i$ is a selection matrix for the $i$th block. Specifically, it is a $1\times 3$ block matrix where the $i$th column is identity, and each block is $6\times 6$. $\boldsymbol{P}^*_1$ selects the first block of a block vector with $6$ blocks of size $6$. Furthermore,  
\begin{align}
    \boldsymbol{\mathcal{J}}_{s, t} &= \boldsymbol{\mathcal{J}}({\ln(\boldsymbol{T}(s, t)\boldsymbol{T}_{n,k}^{-1})}^\vee), \\
    \boldsymbol{\mathcal{T}}_{s, t} &= \text{Ad}(\boldsymbol{T}(s, t)\boldsymbol{T}_{n,k}^{-1}), \\
    \frac{\partial \boldsymbol{\gamma}^\square_{n,k}}{\partial \boldsymbol{x}^\square_{n,k}} &= \begin{bmatrix}
        \boldsymbol{J}_1 & \boldsymbol{0} \\
        \boldsymbol{J}_2 & \boldsymbol{J}_3
    \end{bmatrix}, \\
    \boldsymbol{J}_1 &= \begin{bmatrix}
        \boldsymbol{0} \\
        & \boldsymbol{I} \\
        && \boldsymbol{I}
    \end{bmatrix}, \\
    \boldsymbol{J}_2 &= \begin{bmatrix}
        -\boldsymbol{\mathcal{J}}_{s,t}^{-1}\boldsymbol{\mathcal{T}}_{s,t} & \boldsymbol{0} & \boldsymbol{0} \\
        -\frac{1}{2}\boldsymbol{\epsilon}_{s, t}^\curlywedge\boldsymbol{\mathcal{J}}_{s,t}^{-1}\boldsymbol{\mathcal{T}}_{s,t} & \boldsymbol{0} & \boldsymbol{0} \\
        -\frac{1}{2}\boldsymbol{\varpi}_{s,t}^\curlywedge\boldsymbol{\mathcal{J}}_{s,t}^{-1}\boldsymbol{\mathcal{T}}_{s,t} & \boldsymbol{0} & \boldsymbol{0}
    \end{bmatrix}, \\
    \boldsymbol{J}_3 &= \begin{bmatrix}
        \boldsymbol{\mathcal{J}}_{s,t}^{-1} & \boldsymbol{0} & \boldsymbol{0} \\
        \frac{1}{2}\boldsymbol{\epsilon}_{s, t}^\curlywedge\boldsymbol{\mathcal{J}}_{s,t}^{-1} & \boldsymbol{\mathcal{J}}_{s,t}^{-1} & \boldsymbol{0} \\
        \frac{1}{2}\boldsymbol{\varpi}_{s,t}^\curlywedge\boldsymbol{\mathcal{J}}_{s,t}^{-1} & \boldsymbol{0} & \boldsymbol{\mathcal{J}}_{s,t}^{-1}
    \end{bmatrix}.
\end{align}
\end{subequations}

  \section{Estimation Parameters}\label{app:parameters}

\begin{center}
\small
\setlength{\tabcolsep}{6pt}
\renewcommand{\arraystretch}{1.15}
\begin{tabular}{l p{0.8\linewidth}}
\hline
Label & Diagonal Values \\
\hline
\multicolumn{2}{c}{Simulation Parameters} \\
\hline
$\boldsymbol{R}_{\text{pose}}$\footnotemark & $1\mathrm{E}{-}5,\,1\mathrm{E}{-}5,\,1\mathrm{E}{-}5,\,2.5\mathrm{E}{-}4,\,2.5\mathrm{E}{-}4,\,2.5\mathrm{E}{-}4$ \\
$\boldsymbol{Q}_0$ & $3.2\mathrm{E}{-}5,\,3.2\mathrm{E}{-}5,\,3.2\mathrm{E}{-}5,\,4\mathrm{E}{-}3,\,4\mathrm{E}{-}3,\,4\mathrm{E}{-}3$, \\
                   & $4\mathrm{E}{-}3,\,4\mathrm{E}{-}3,4\mathrm{E}3,\,4\mathrm{E}3,\,4\mathrm{E}3,\,4\mathrm{E}{-}3$, \\
                   & $4\mathrm{E}{-}3,\,4\mathrm{E}{-}3,\,4\mathrm{E}{-}3,\,4\mathrm{E}{-}3,\,4\mathrm{E}{-}3,\,4\mathrm{E}{-}3$ \\
$\boldsymbol{Q}_1$ & $2.8\mathrm{E}{-}2,\,2.8\mathrm{E}{-}2,\,2.8\mathrm{E}{-}2,\,8\mathrm{E}{-}2,\,8\mathrm{E}{-}2,\,8\mathrm{E}{-}2$ \\
$\boldsymbol{Q}_2$ & $4\mathrm{E}{-}1,\,4\mathrm{E}{-}1,\,4\mathrm{E}{-}1,\,4\mathrm{E}2,\,4\mathrm{E}2,\,4\mathrm{E}2$ \\
$\boldsymbol{Q}_3$ & $4\mathrm{E}0,\,4\mathrm{E}0,\,4\mathrm{E}2,\,1.6\mathrm{E}3,\,1.6\mathrm{E}3,\,4\mathrm{E}0$ \\
$\boldsymbol{Q}_4$ & $4\mathrm{E}0,\,4\mathrm{E}0,\,4\mathrm{E}0,\,4\mathrm{E}1,\,4\mathrm{E}1,\,4\mathrm{E}1$ \\
\hline
\multicolumn{2}{c}{Experiment Parameters} \\
\hline
$\boldsymbol{R}_{\text{pose}}$ & $3\mathrm{E}{-}5,\,3\mathrm{E}{-}5,\,3\mathrm{E}{-}5,\,2.5\mathrm{E}{-}3,\,2.5\mathrm{E}{-}3,\,2.5\mathrm{E}{-}3$ \\
$\boldsymbol{R}_{\text{gyro}}$ & $1\mathrm{E}{-}4,\,1\mathrm{E}{-}4,\,1\mathrm{E}{-}4$ \\
$\boldsymbol{Q}_0$ & $4\mathrm{E}{-}5,\,4\mathrm{E}{-}5,\,4\mathrm{E}{-}5,\,5\mathrm{E}{-}3,\,5\mathrm{E}{-}3,\,5\mathrm{E}{-}3$, \\
                   & $5\mathrm{E}{-}3,\,5\mathrm{E}{-}3,\,5\mathrm{E}{-}3,\,5\mathrm{E}1,\,5\mathrm{E}1,\,5\mathrm{E}{-}3$, \\
                   & $5\mathrm{E}{-}3,\,5\mathrm{E}{-}3,\,5\mathrm{E}{-}3,\,5\mathrm{E}{-}3,\,5\mathrm{E}{-}3,\,5\mathrm{E}{-}3$ \\
$\boldsymbol{Q}_1$ & $1.5\mathrm{E}0,\,1.5\mathrm{E}0,\,1.5\mathrm{E}0,\,1.5\mathrm{E}2,\,1.5\mathrm{E}2,\,1.5\mathrm{E}2$ \\
$\boldsymbol{Q}_2$ & $4\mathrm{E}{-}1,\,4\mathrm{E}{-}1,\,4\mathrm{E}{-}1,\,2.5\mathrm{E}2,\,2.5\mathrm{E}2,\,2.5\mathrm{E}2$ \\
$\boldsymbol{Q}_3$ & $1.5\mathrm{E}1,\,1.5\mathrm{E}1,\,1.5\mathrm{E}1,\,1.5\mathrm{E}2,\,1.5\mathrm{E}2,\,1.5\mathrm{E}2$ \\
$\boldsymbol{Q}_4$ & $1.5\mathrm{E}1,\,1.5\mathrm{E}1,\,1.5\mathrm{E}1,\,2.5\mathrm{E}1,\,2.5\mathrm{E}1,\,2.5\mathrm{E}1$ \\
\hline
\end{tabular}
\end{center}

\footnotetext{Note that $\boldsymbol{R}_{\text{pose}}$ is doubled for the `Simulation High Noise' trial.}

\end{appendices}

\bibliographystyle{IEEEtran}
\bibliography{IEEEabrv,refs}

\begin{IEEEbiography}[{\includegraphics[width=1in,height=1.25in,clip,keepaspectratio]{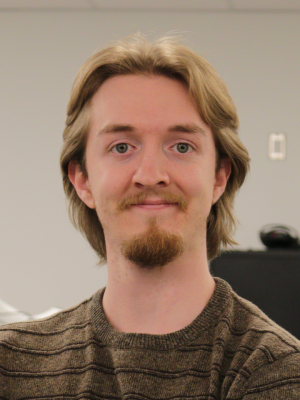}}]{Spencer Teetaert}
  (Student Member, IEEE) received a B.A.Sc.\ degree in Engineering Science in 2023 from the University of Toronto, ON, Canada. He is currently a Ph.D.\ candidate with the Robotics Institute at the University of Toronto. His research focus is on stochastic state estimation for continuum robotics. 
\end{IEEEbiography}

\begin{IEEEbiography}[{\includegraphics[width=1in,height=1.25in,clip,keepaspectratio]{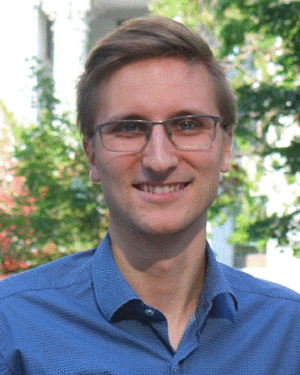}}]{Sven Lilge} (Member, IEEE) received the B.Sc.\ and M.Sc.\ degrees in electrical engineering and information technology from Leibniz University Hannover, Hanover, Germany, in 2014 and 2016, respectively, and the Ph.D. degree in computer science from the University of Toronto, Toronto, ON, Canada, in 2023. He is currently an Assistant Professor with the Department of Computer Science, Toronto Metropolitan University, Toronto, ON, Canada, where he directs the Autonomous Robotics and Continuum Systems Lab. Previously, he was a Postdoctoral Fellow with the University of Toronto Robotics Institute, Toronto. His research focuses on fundamental challenges in algorithmic autonomy for continuum and mobile robots, specifically advancing motion planning, control, state estimation, and human-robot interaction in complex environments.
\end{IEEEbiography}

\begin{IEEEbiography}[{\includegraphics[width=1in,height=1.25in,clip,keepaspectratio]{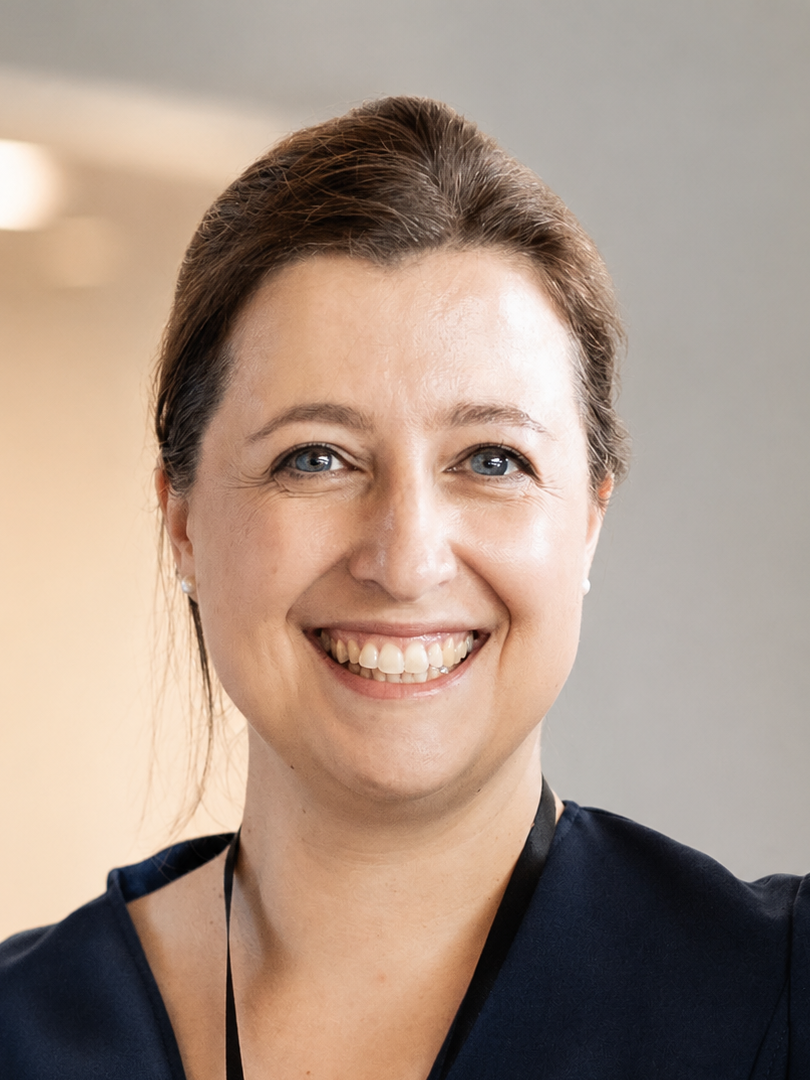}}]{Jessica Burgner-Kahrs} (Senior Member, IEEE) received the diploma and the Ph.D. degree in computer science from the Karlsruhe Institute of Technology (KIT), Karlsruhe, Germany, in 2006 and 2010, respectively. She is a Professor with the Departments for Mathematical and Computational Sciences, Computer Science, and Mechanical and Industrial Engineering, and the founding Director of the Continuum Robotics Laboratory, University of Toronto, Toronto, ON, Canada. From 2013 to 2019, she was with Leibniz University Hannover, Hanover, Germany, and from 2010 to 2012, with Vanderbilt University, Nashville, TN, USA\@. Her fundamental robotics research is driven by applications in minimally invasive surgery and maintenance, repair, and operations. Her research interests include continuum robotics and in particular on their design, modeling, planning, and control, as well as human-robot interaction.
\end{IEEEbiography}

\begin{IEEEbiography}[{\includegraphics[width=1in,height=1.25in,clip,keepaspectratio]{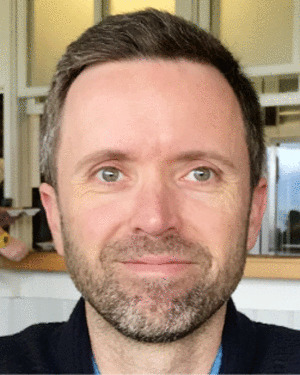}}]{Timothy D. Barfoot} (Fellow, IEEE) received the B.A.Sc.\ degree in engineering science and the Ph.D. degree in aerospace science and engineering from the University of Toronto, Toronto, ON, Canada, in 1997 and 2002, respectively. He is currently a Professor with the University of Toronto Robotics Institute, Toronto. His research focuses on enabling robotic systems to operate robustly over long durations in large-scale, unstructured, 3-D environments using rich onboard sensing and computation. He has authored State Estimation for Robotics (Cambridge, 2017, 2024) and coauthored the SLAM Handbook (Cambridge, 2026). His research interests include guidance, navigation, and control of autonomous systems for a wide range of applications. Dr.~Barfoot is a Fellow of the Canadian Academy of Engineering. He is currently the Editor-in-Chief of IEEE Transactions on Field Robotics.
\end{IEEEbiography}

\end{document}